\documentclass[letterpaper, 10 pt, conference]{ieeeconf}  % Comment this line out if you need a4paper

\IEEEoverridecommandlockouts                              % This command is only needed if 
\usepackage{epsfig} % for postscript graphics files
\usepackage{times} % assumes new font selection scheme installed
\usepackage{amsmath} % assumes amsmath package installed
\usepackage{amssymb}  % assumes amsmath package installed
\usepackage{graphicx}
\usepackage{caption}
\usepackage{hyperref}
\usepackage{tikz}
\usetikzlibrary{shapes.geometric, arrows}
\usetikzlibrary{3d,fit, positioning}

\usepackage{import}
\usetikzlibrary{quotes,arrows.meta}

\usepackage{Box}

\graphicspath{ {./images/}}

\begin{document}

\title{\LARGE \bf Unsupervised Traffic Scene Generation with Synthetic 3D Scene Graphs}

\author{Artem Savkin$^{1,2}$\\TUM, BMW \and Rachid Ellouze$^{1,2}$\\TUM, BMW \and Nassir Navab$^{1,3}$\\TUM, JHU \and Federico Tombari$^{1,4}$\\TUM, Google% <-this % stops a space
%\thanks{*Equal Contribution}% <-this % stops a space
\thanks{$^{1}$TUM, Boltzmannstr. 3, 85748 Munich, Germany
{\tt\small artem.savkin@tum.de};
%{\tt\small rachid.ellouze@tum.de};
{\tt\small tombari@in.tum.de}}%
\thanks{$^{2}$BMW AG, 80809 Munich, Germany}
\thanks{$^{1}$JHU, Baltimore, MD 21218, United States}
\thanks{$^{4}$Google, 8002 Zurich, Switzerland}
}

%\author{Anonymous submission}

\newcommand\blfootnote[1]{%
	\begingroup
	\renewcommand\thefootnote{}\footnote{#1}%
	\addtocounter{footnote}{-1}%
	\endgroup
}

\newcommand{\norm}[1]{\left\lVert#1\right\rVert}
\newcommand{\TODO}[1] {\textbf{[TODO: #1]}}
\newcommand{\Loss}{\mathcal{L}}
\newcommand{\Exp}{\mathop{\mathbb{E}}}
\newcommand{\rarr}{\rightarrow}
\newcommand{\larr}{\leftarrow}

\maketitle

%\twocolumn[{%
%\renewcommand\twocolumn[1][]{#1}%
%\maketitle
%\input{figure_teaser}
%}]

\begin{abstract}
Image synthesis driven by computer graphics achieved recently a remarkable realism, yet synthetic image data generated this way reveals a significant domain gap with respect to real-world data. This is especially true in autonomous driving scenarios, which represent a critical aspect for overcoming utilizing synthetic data for training neural networks. We propose a method based on domain-invariant scene representation to directly synthesize traffic scene imagery without rendering. Specifically, we rely on synthetic scene graphs as our internal representation and introduce an unsupervised neural network architecture for realistic traffic scene synthesis. We enhance synthetic scene graphs with spatial information about the scene and demonstrate the effectiveness of our approach through scene manipulation.
\end{abstract}

%\blfootnote{%\thanks{*Equal Contribution}% <-this % stops a space
%$^{1}$TUM, 85748 Munich, Germany
%{\tt\small artem.savkin@tum.de};
%{\tt\small rachid.ellouze@tum.de};
%{\tt\small tombari@in.tum.de}
%}%
%\blfootnote{$^{2}$BMW AG, 80809 Munich (Germany)}
%\blfootnote{$^{3}$JHU, Baltimore, MD 21218, United States}
%\blfootnote{$^{4}$Google, 8002 Zurich, Switzerland}

%\addtolength{\textheight}{-12cm}
% the text height of the last page by a suitable amount.

\section{Introduction}

A broad variety of real world scenarios require autonomous navigation systems to rely on machine learning-based perception algorithms. Such algorithms are knowingly data dependent, yet data acquisition and labeling is a costly and tedious process. It is associated with manual labor, must handle rare \textit{long tail} corner case events, and could be hard constrained by ethical aspects e.g., in case of near-accident scenarios.

%\begin{center}
\begin{figure}[!ht]

\centering
% \includegraphics[width=0.325\textwidth]{images/pfd2bdd/graphs/74.png}
% \includegraphics[width=0.325\textwidth]{images/pfd2bdd/images/74.png}
% \includegraphics[width=0.325\textwidth]{images/pfd2cs/images/74.png}
% \captionof{figure}{Synthetic scene graph and generated traffic scenes from BDD (left) \cite{Yu2018} and Cityscapes (right) \cite{Cordts2016} datasets.}

\begin{minipage}{0.95\columnwidth}
\centering
\includegraphics[width=1\textwidth]{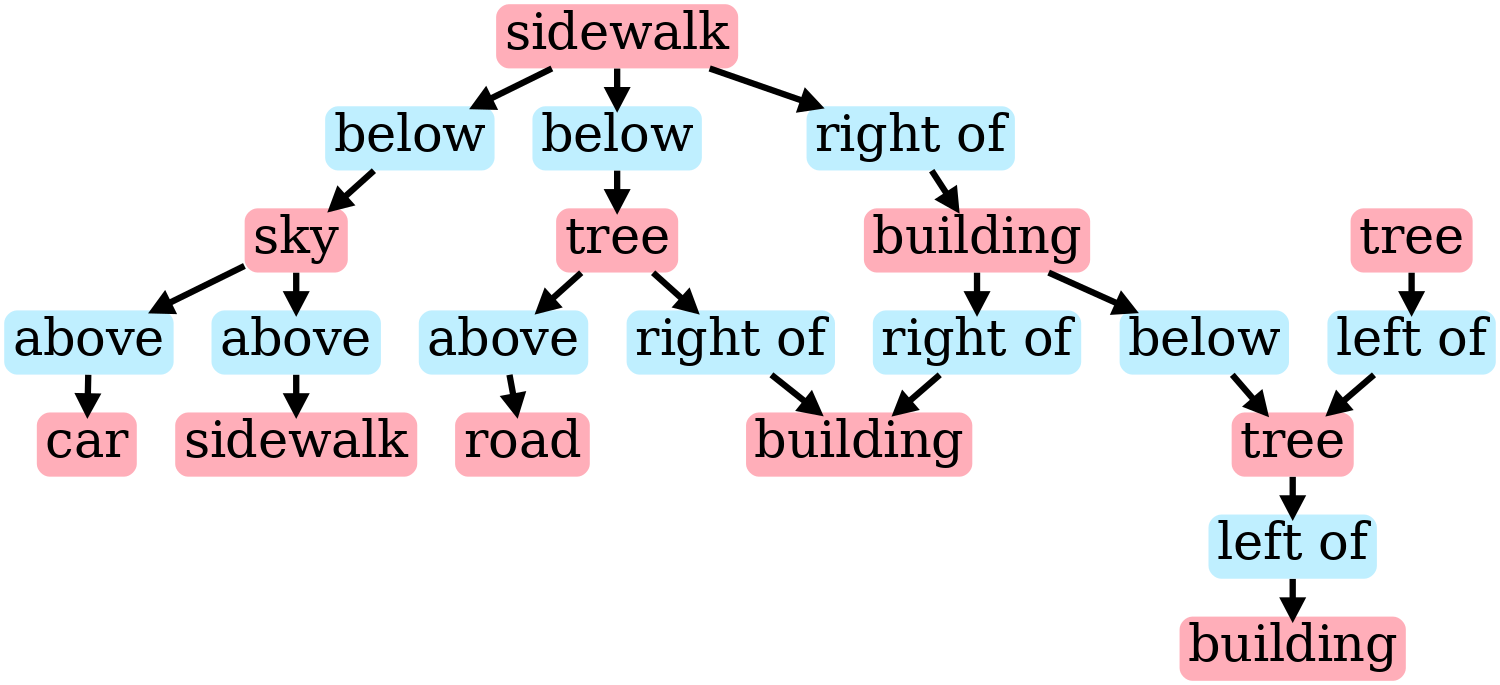}
\includegraphics[width=1\textwidth]{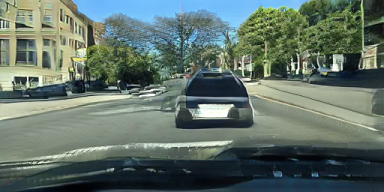}
\includegraphics[width=1\textwidth]{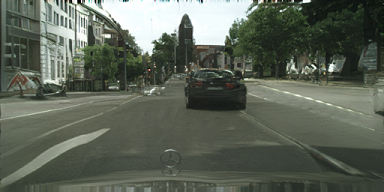}
% \caption*{Original}
\end{minipage}
%\begin{minipage}{0.325\textwidth}
%\centering
%\includegraphics[height=0.5\textwidth]{images/pfd2bdd/graphs/562.png}
%\includegraphics[width=1\textwidth]{images/pfd2bdd/images/562.png}
%\includegraphics[width=1\textwidth]{images/pfd2cs/images/562.png}
%% \caption*{Original}
%\end{minipage}
%\begin{minipage}{0.325\textwidth}
%\centering
%includegraphics[height=0.5\textwidth]{images/pfd2bdd/graphs/88.png}
%\includegraphics[height=0.5\textwidth]{images/pfd2bdd/images/88.png}
%\includegraphics[height=0.5\textwidth]{images/pfd2cs/images/88.png}
%% \caption*{Original}
%\end{minipage}
\caption{BDD and Cityscapes traffic scenes generated from synthetic scene graph.}
\label{fig:teaser}
%\end{center}
\end{figure}

One of the common alternatives to real data acquisition and annotation is represented by simulation and synthetic data. Simulation has a long history in driver assistance systems, but with the renaissance of neural networks, the research community has strengthened efforts in this direction and many synthetic datasets \cite{Ros2016, Richter2017, Wrenninge2020} and simulation systems \cite{Dosovitskiy2017, Shah2017} appeared.

\begin{figure*}[t!]
\centering
\begin{tikzpicture}

\definecolor{color_road}{RGB}{84,0,84}
\definecolor{color_car}{RGB}{0,0,148}
\definecolor{color_sidewalk}{RGB}{254,36,242}
\definecolor{color_building}{RGB}{73,73,73}
\definecolor{color_sky}{RGB}{73,135,188}

\def\colorLayout{red!5}
\def\colorImageBlue{rgb:blue,5;green,2.5;white,5}
\def\colorImageDark{green!50}
\def\colorImage{green!5}

\tikzstyle{node} = [rectangle, minimum size=1cm, text centered, draw=black!50]
\tikzstyle{sg} = [rectangle, rounded corners, fill=black!0]
\tikzstyle{layer} = [rectangle, minimum size=20mm, draw=black!20]
\tikzstyle{encoder} = [trapezium, text centered, draw=black]
\tikzstyle{vector} = [rectangle, minimum width=0.5cm, text centered]
% \tikzstyle{mask} = [trapezium, text centered, draw=black]
\tikzstyle{mask} = [matrix, text centered]

\node(node2) [sg, xshift=0cm]
{\includegraphics[width=20mm]{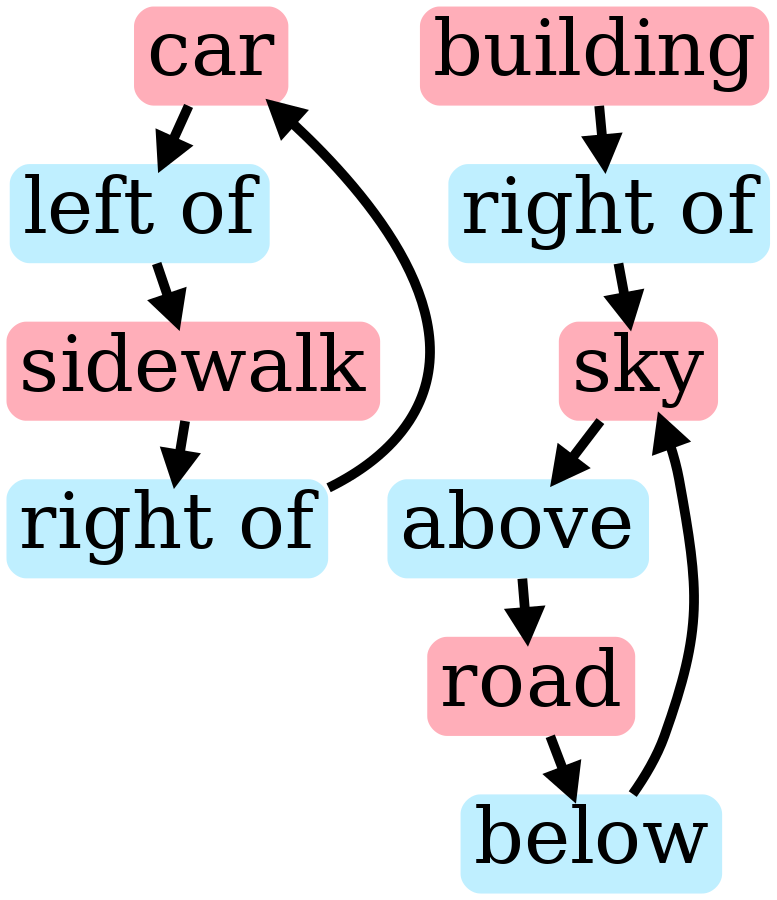}};

% \node(node1) [node, xshift=0cm, yshift=-3.0cm]
% {\includegraphics[width=20mm]{images/render.png}};
\node(node1) [canvas is zy plane at x=0.0, yshift=-3cm]
{\includegraphics[width=30mm]{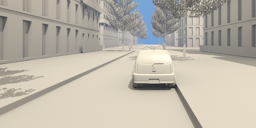}};
%{\includegraphics[width=30mm]{images/01743.png}};

\pic[shift={(2.25,0,0)}] at (node2) {Box={name=enc11, caption=, fill=gray, opacity=0.5,height=8,width=1,depth=8}};
\pic[shift={(0.25,0,0)}] at (enc11-east) {Box={name=enc12, caption=, fill=gray, opacity=0.5,height=6,width=1,depth=6}};

\pic[shift={(0.5,0,0)}] at (enc12-east) {Box={name=z1, caption=, fill=color_sidewalk, opacity=0.9, draw=color_sidewalk, height=1,width=3,depth=1}};
\pic[shift={(0.5,0.4,0)}] at (enc12-east) {Box={name=z2, caption=, fill=color_building, opacity=0.9, height=1,width=3,depth=1}};
\pic[shift={(0.5,-0.4,0)}] at (enc12-east) {Box={name=z3, caption=, fill=color_car, opacity=0.9, height=1, width=3, depth=1}};
\pic[shift={(0.5,0.8,0)}] at (enc12-east) {Box={name=z4, caption=, fill=color_sky, opacity=0.9, height=1, width=3, depth=1}};
\pic[shift={(0.5,-0.8,0)}] at (enc12-east) {Box={name=z4, caption=, fill=color_road, opacity=0.9, height=1, width=3, depth=1}};

%DECODER 1

\pic[shift={(1.5,0,0)}] at (enc12-east) {Box={name=dec11, caption=, fill=\colorImageBlue,opacity=0.5,height=6,width=1,depth=6}};
\pic[shift={(0.25,0,0)}] at (dec11-east) {Box={name=dec12, caption=, fill=\colorImageBlue,opacity=0.5,height=8,width=1,depth=8}};

%LOSSES

\node(node80) [rectangle, fill=\colorImageBlue, opacity=0.1, rounded corners, minimum height=6.5cm, minimum width=3cm, xshift=7.5cm, yshift=-1.5cm] {};

\node(node81) [canvas is zy plane at x=7.5, yshift=-3cm]
{\includegraphics[width=30mm]{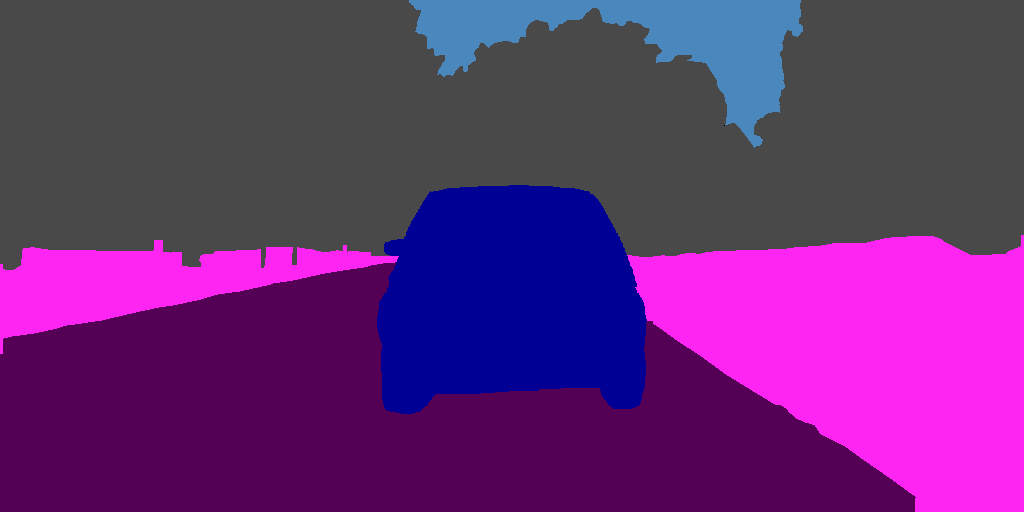}};

%OUTPUT 1

\node(node9) [layer, canvas is zy plane at x=7.0]
{\includegraphics[width=30mm]{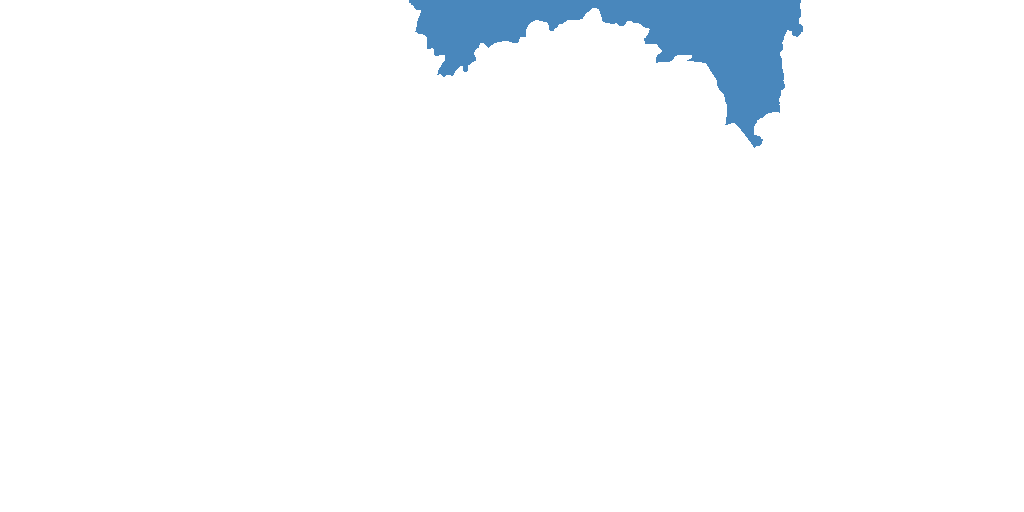}};
\node(node10) [layer, canvas is zy plane at x=7.25]
{\includegraphics[width=30mm]{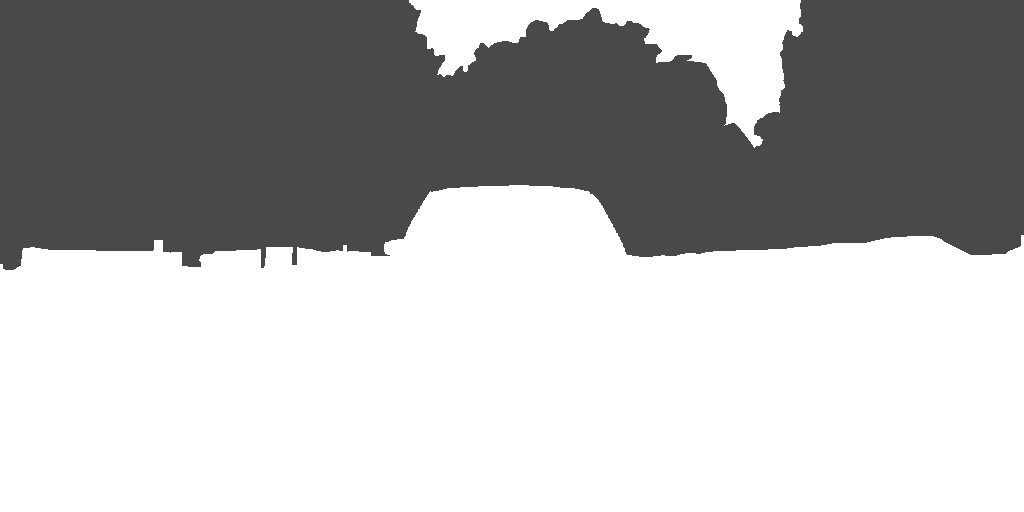}};
\node(node11) [layer, canvas is zy plane at x=7.5]
{\includegraphics[width=30mm]{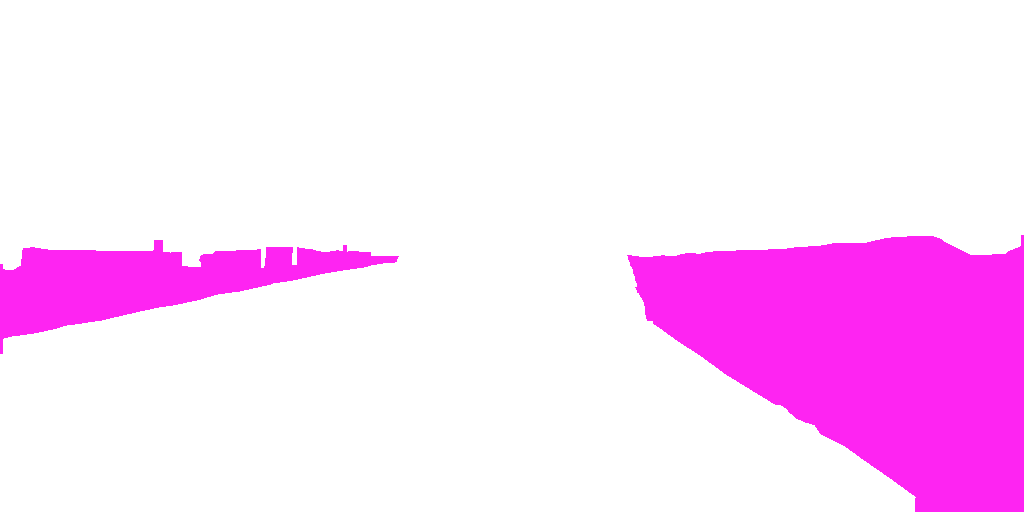}};
\node(node12) [layer, canvas is zy plane at x=7.75]
{\includegraphics[width=30mm]{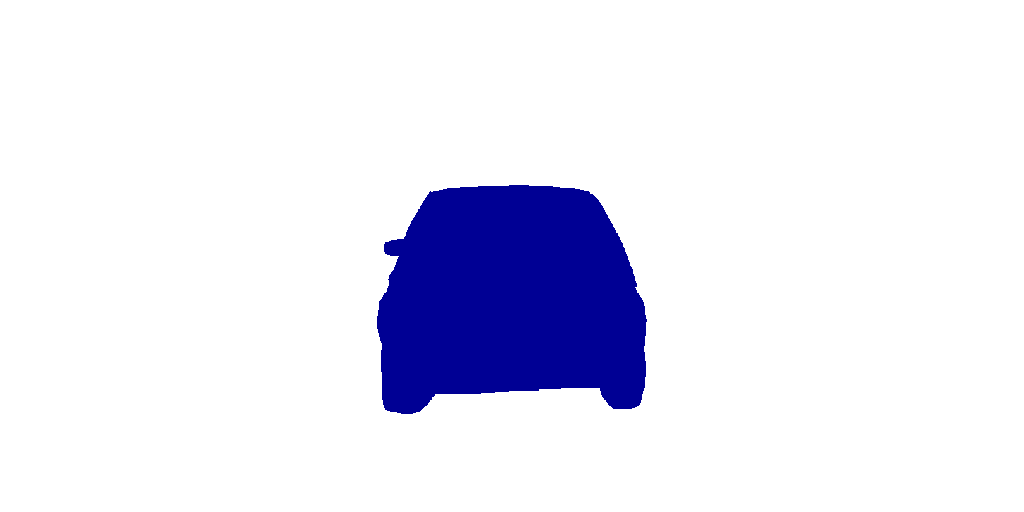}};
\node(node13) [layer, canvas is zy plane at x=8.0]
{\includegraphics[width=30mm]{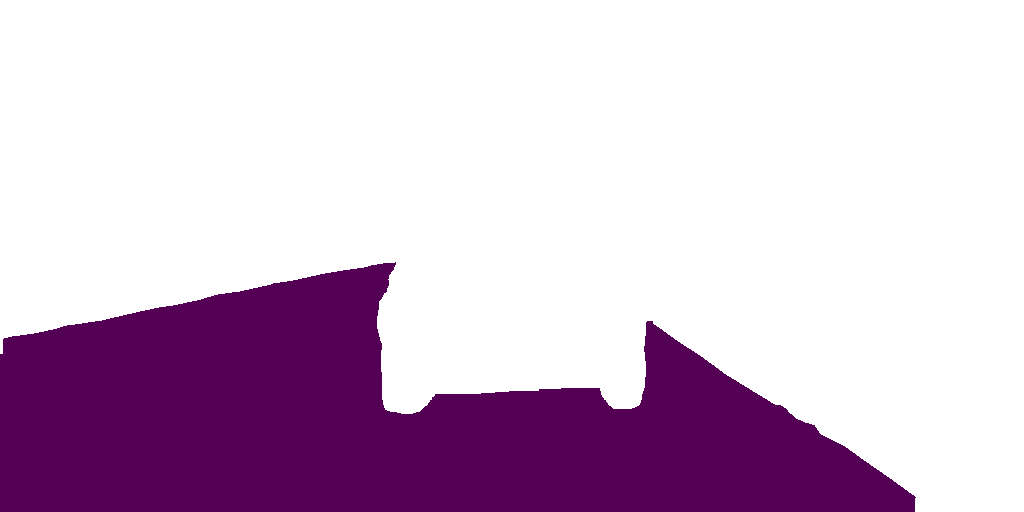}};

%NETWORK 2

\pic[shift={(2.25,0,0)}] at (node13) {Box={name=enc21, caption=, fill=\colorImageBlue,opacity=0.5,height=8,width=1,depth=8}};
\pic[shift={(0.25,0,0)}] at (enc21-east) {Box={name=enc22, caption=, fill=\colorImageBlue,opacity=0.5,height=6,width=1,depth=6}};
\pic[shift={(0.25,0,0)}] at (enc22-east) {Box={name=enc23, caption=, fill=\colorImageBlue,opacity=0.5,height=4,width=1,depth=4}};

\pic[shift={(0.25,0,0)}] at (enc23-east) {Box={name=dec21, caption=, fill=\colorImageDark,opacity=0.5,height=6,width=1,depth=6}};
\pic[shift={(0.25,0,0)}] at (dec21-east) {Box={name=dec22, caption=, fill=\colorImageDark, height=8,width=1,depth=8}};

%LOSS 2

\node(node160) [rectangle, fill=\colorImage, rounded corners, minimum height=6.5cm, minimum width=3cm, xshift=15.0cm, yshift=-1.5cm] {};

\node(node15) [canvas is zy plane at x=15.0]
{\includegraphics[width=30mm]{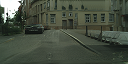}};

\node(node16) [canvas is zy plane at x=15.0, yshift=-3cm]
{\includegraphics[width=30mm]{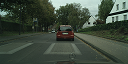}};

% EDGES

\draw[->, draw=black!50] (node1.north) -- (node2.south);

\draw[->, draw=black!50] ([xshift=0.75cm]node1.center) -- ([xshift=-0.75cm]node81.center);
\draw[->, draw=black!50] (node2.east) -- ([xshift=-0.5cm]enc11-west);

\draw[->, draw=black!50] (dec12-east) -- ([xshift=-0.75cm]node9.center);

\draw[->, draw=black!50] ([xshift=0.75cm]node13.center) -- ([xshift=-0.5cm]enc21-west);
\draw[->, draw=black!50] (dec22-east) -- ([xshift=-0.75cm]node15.center);

% \node(n_i)[xshift=1.0cm, yshift=-1.5cm]{$\{n_i,e_{i,j}\}$};

\node(P)[xshift=3.0cm, yshift=-1.5cm]{$P$};
\node(G)[xshift=11.0cm, yshift=-1.5cm]{$G$};

% \node(x)[xshift=14.5cm, yshift=1.0cm]{$x$};

\node(G)[xshift=8.5cm, yshift=-4.5cm]{$\mathcal{L}_{SG}$};
\node(G)[xshift=16.0cm, yshift=-4.5cm]{$\mathcal{L}_{TS}$};

\node(x)[xshift=6.5cm,yshift=1.0cm]{$b_i, m_i$};
\node(x)[xshift=6.5cm,yshift=-2.5cm]{$\hat{b}_i, \hat{m}_i$};
\node(x)[xshift=14.0cm,yshift=0.5cm]{$x$};
\node(x)[xshift=14.0cm,yshift=-2.5cm]{$\hat{x}$};

\end{tikzpicture}
\caption{Overview of the network with graph processor $P$, traffic scene generator $G$, and loss functions $\mathcal{L}_{SG}$ and $\mathcal{L}_{TS}$.}\label{fig:scheme}
\end{figure*}
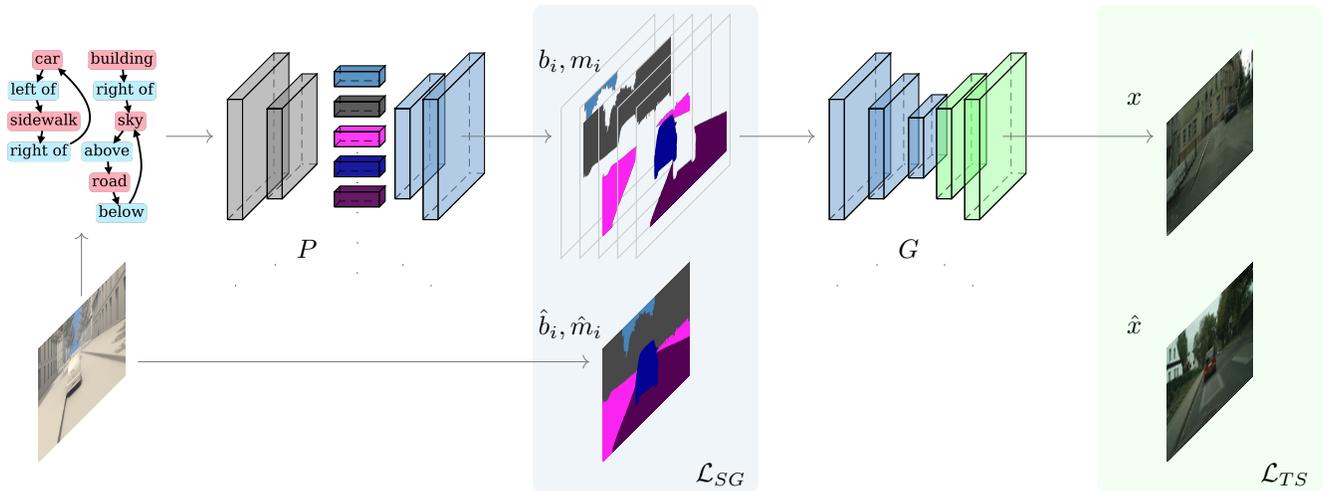

Despite being a powerful research tool, synthetic data typically reveals a significant domain gap with respect to target real data. The underlying phenomenon, where marginal distributions in both domains differ, has been defined by \cite{Sugiyama2012} as a \textit{covariate shift}. This means that \textit{i.i.d} assumption does not hold for a synthetic-real setup. Simply stated, synthetic and real domains differ both in content and appearance. This problem is normally a subject for domain adaptation methods such as \textit{sim2real} domain transfer.

In this work, we propose overcoming the domain gap between both domains by eliminating the rendering part from the pipeline. The intuition behind this concept tells that rendered images introduce a bias towards the underlying domain, which is then commonly leveled out by e.g., style transfer methods. Our main idea is to replace the rendering with abstract scene representation and directly synthesize realistic images of traffic scenes from it. We utilize scene graphs as such abstract representations. Scene graphs encode objects in the scene as nodes and relations between them as edges \cite{Johnson2015}, in addition, they can also integrate certain characteristics of the objects as attributes \cite{Ashual2019}.

In this regard, scene graphs are domain agnostic as they can be generated in simulation and be applied to real-world data without a strong domain gap. In fact, scene graphs are fairly simple to simulate and manipulate, which allows for domain randomization and data generation of potentially arbitrary size and variance.

Derived from a simulation, synthetic scene graphs incorporate relevant objects as \textit{car, person} etc. and relations like \textit{left to}, \textit{right to}. We additionally extend this setup with necessary traffic scene classes like \textit{road, sidewalk, building, vegetation}. It is important to note that simulation provides 3D information about the scene, which can be introduced into the scene graph in the form of spatial attributes and spatial relations between objects. Moreover, synthetic annotations provided by simulation are pixel precise and could be used in a downstream task.

In this work, we propose synthetic 3D scene graphs with spatial components. We also provide the aforementioned graph representations for existing synthetic urban traffic datasets PfB \cite{Richter2017} and Synscapes \cite{Wrenninge2020}. To enable realistic traffic scene generation from synthetic scene graphs, we propose a neural network architecture that supports unsupervised training. We demonstrate the benefit of our approach with an online tool through traffic scene generation and manipulation: \href{https://artemsavkin.github.io/sg2ts/}{Demo}.
\section{Related Work}

\textit{Rendering.}
Historically, traffic scene synthesis was commonly achieved through computer graphics. Multiple works focused on physically realistic rendering of urban traffic environments: SYNTHIA \cite{Ros2016}, Virtual KITTI \cite{Gaidon2016}, PfB \cite{Richter2017}, VKITTI2 \cite{Cabon2020}. The most recent and visually realistic one is arguably Synscapes \cite{Wrenninge2020}. Simulation can be applied not only in outdoor environments, but also in indoor scene synthesis for autonomous agents \cite{McCormac2017}, \cite{Qiu2017}. Although rendered data reveals a high level of realism and a great deal of variations, it is still affected by a significant domain gap with respect to real data when it is used for training machine learning approaches, in particular, neural networks.\\
\textit{Domain Transfer.}
To mitigate the limitation introduced by \textit{sim2real} domain shift, research has focused on data synthesis using deep neural networks. The vast majority of such models \cite{Odena2017}, \cite{Gulrajani2017}, \cite{Brock2018}, \cite{Karras2019}, utilizes adversarial frameworks (GAN) \cite{Goodfellow2014} for image generation. Many of them condition the generation process on visual artifacts or descriptions derived from available labels (semantics, edges etc.), such as Pix2Pix \cite{Isola2017}, CRN \cite{Chen2017}, Pix2PixHD \cite{Wang2018} and SPADE \cite{Park2019}. There are also several works which employed unsupervised domain adaptation from synthetic to real images, such as CycleGAN \cite{Zhu2017, Almahairi2018}, DIRT-T \cite{Shu2018}, MUNIT \cite{Huang2018}.\\
\textit{Scene Graphs.} Another relevant research direction is represented by the use of more domain invariant scene descriptions such as text \cite{Zhang2017} or scene graphs. Scene graphs are abstract data structures used for describing scenes by encoding objects as nodes and relations between them as edges. Thus, the whole scene can be represented as a directed graph. Scene graphs have been used as an alternative to natural language description for image retrieval \cite{Johnson2015} and image description \cite{Newell2017}. Most recent works on scene graphs for image generation include \cite{Johnson2018}, \cite{Ashual2019} and \cite{Dhamo2020}. \cite{Ashual2019} proposed dual layout and appearance embedding for better matching of generated scenes and underlying scene graphs. \cite{Dhamo2020} utilizes scene graphs as an intermediate representation for image manipulation without direct supervision. A lot of works in the area of scene graphs rely on the Visual Genome dataset \cite{Krishna2017}, which provides image samples annotated with scene graphs. To our knowledge, there is no such dataset available for urban traffic environments, so we provide scene graph annotations for commonly used public datasets, in particular, PfB \cite{Richter2017} and Synscapes \cite{Wrenninge2020}.
\begin{figure*}[ht]

\begin{minipage}{0.195\textwidth}
\centering
\includegraphics[height=1\textwidth]{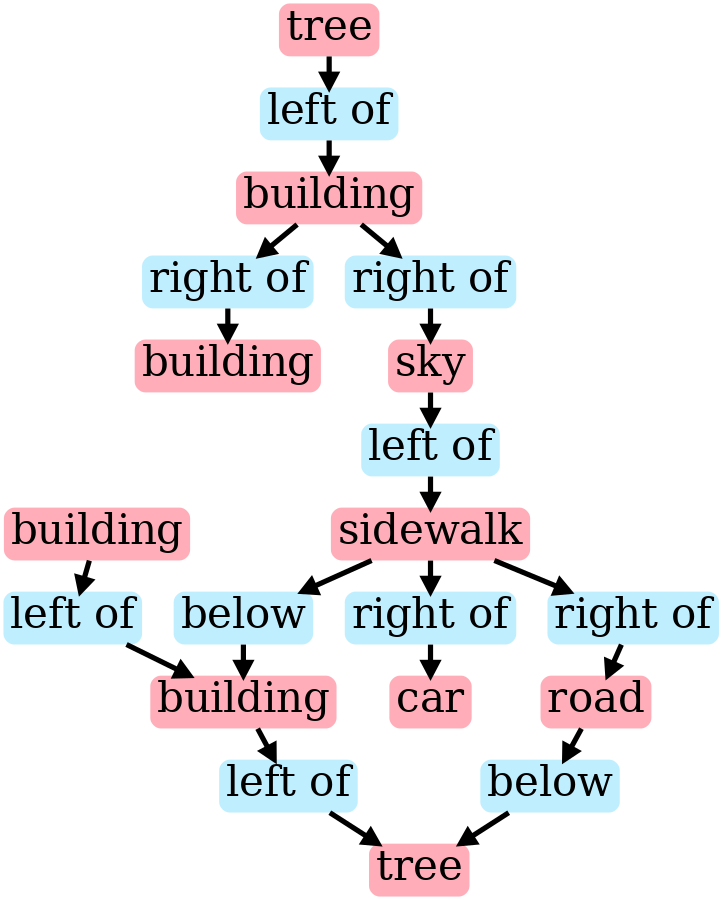}
\includegraphics[width=1\textwidth]{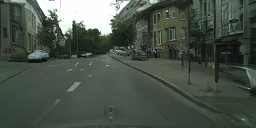}
\includegraphics[width=1\textwidth]{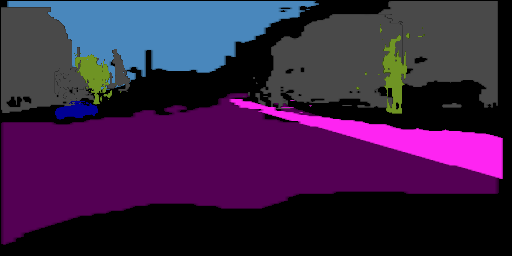}
\end{minipage}
\vspace{0.005\textwidth}
% \begin{minipage}{0.195\textwidth}
% \centering
% \includegraphics[height=1\textwidth]{images/pfd2cs/graphs/26.png}
% \includegraphics[width=1\textwidth]{images/pfd2cs/layouts/26.png}
% \includegraphics[width=1\textwidth]{images/pfd2cs/images/26.png}
% \end{minipage}
% \begin{minipage}{0.195\textwidth}
% \centering
% \includegraphics[height=1\textwidth]{images/pfd2cs/graphs/29.png}
% \includegraphics[width=1\textwidth]{images/pfd2cs/layouts/29.png}
% \includegraphics[width=1\textwidth]{images/pfd2cs/images/29.png}
% \end{minipage}
\vspace{0.005\textwidth}
\begin{minipage}{0.195\textwidth}
\centering
\includegraphics[height=1\textwidth]{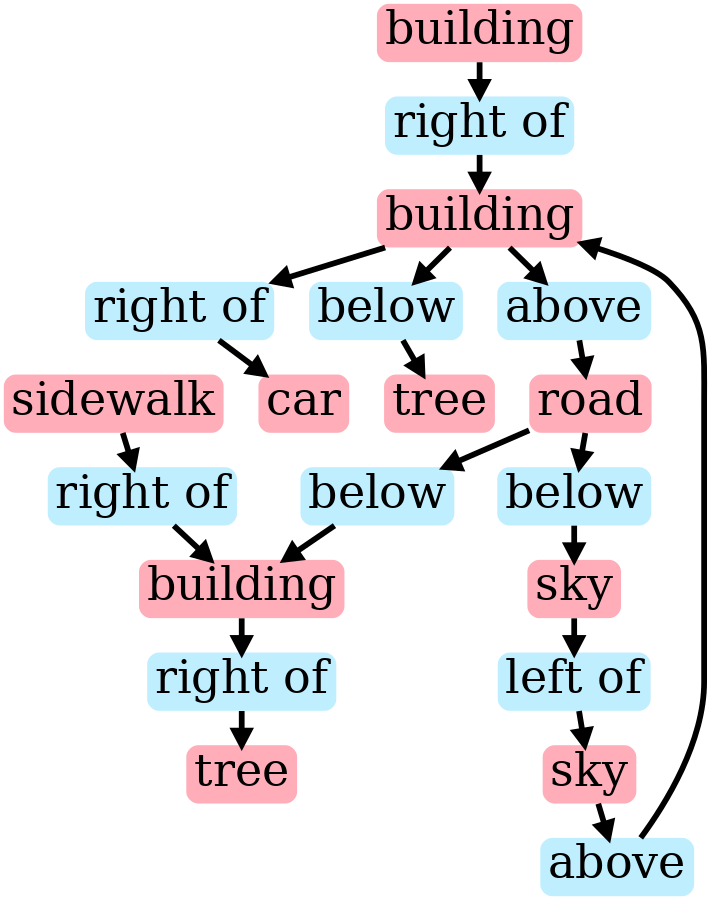}
\includegraphics[width=1\textwidth]{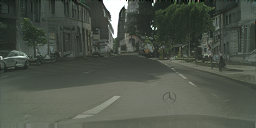}
\includegraphics[width=1\textwidth]{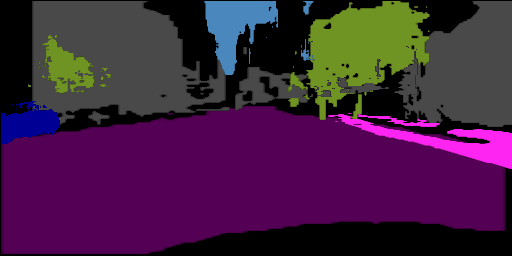}
\end{minipage}
\begin{minipage}{0.195\textwidth}
\centering
\includegraphics[height=1\textwidth]{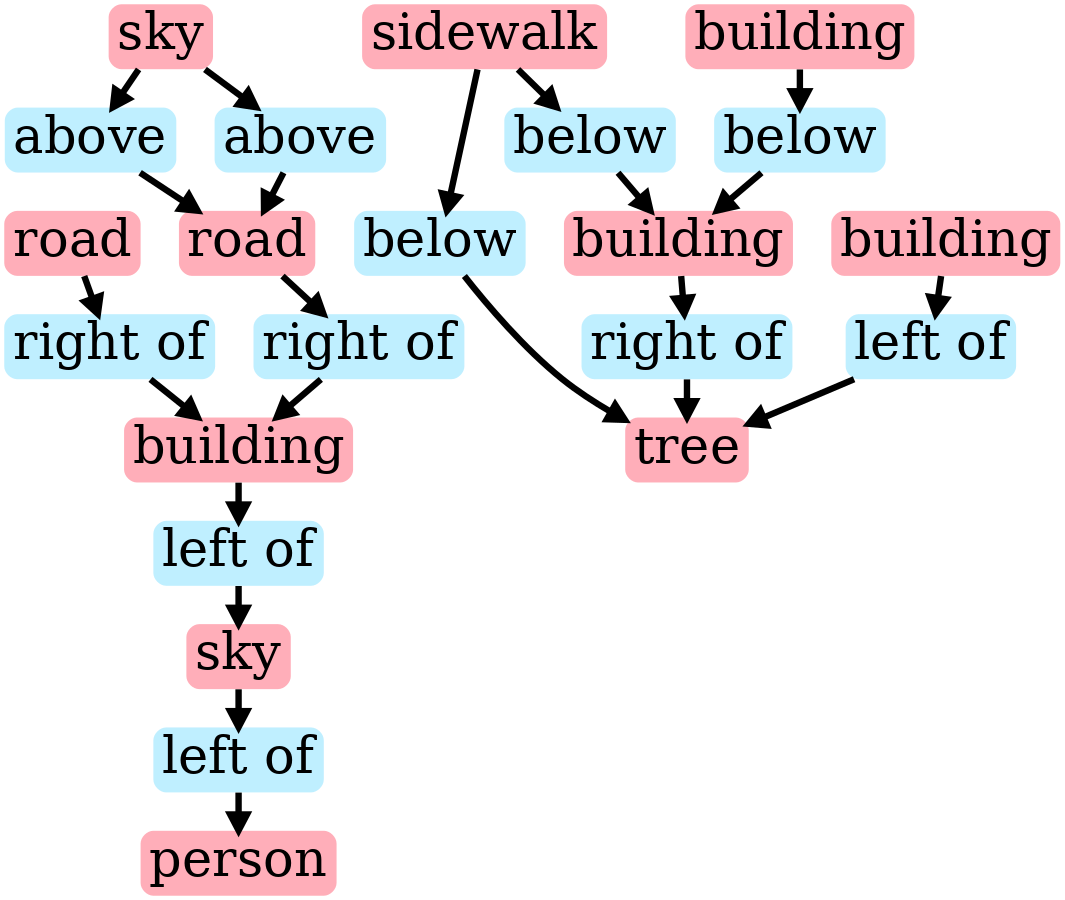}
\includegraphics[width=1\textwidth]{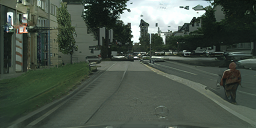}
\includegraphics[width=1\textwidth]{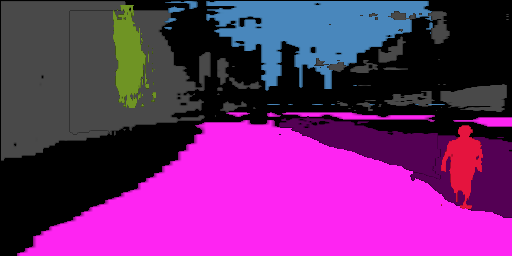}
\end{minipage}
\vspace{0.005\textwidth}
\begin{minipage}{0.195\textwidth}
\centering
\includegraphics[height=1\textwidth]{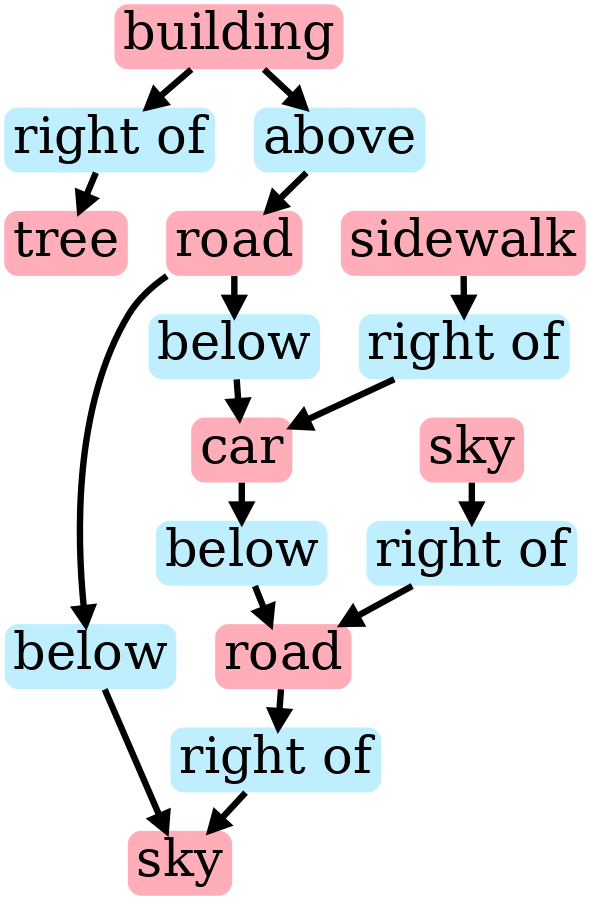}
\includegraphics[width=1\textwidth]{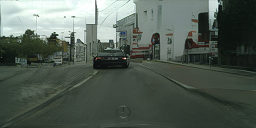}
\includegraphics[width=1\textwidth]{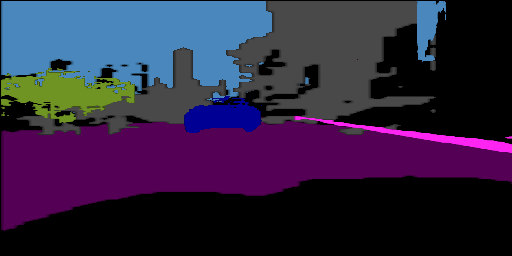}
\end{minipage}
\vspace{0.005\textwidth}
% \begin{minipage}{0.195\textwidth}
% \centering
% \includegraphics[height=1\textwidth]{images/pfd2cs/graphs/67.png}
% \includegraphics[width=1\textwidth]{images/pfd2cs/layouts/67.png}
% \includegraphics[width=1\textwidth]{images/pfd2cs/images/67.png}
% \end{minipage}
% \begin{minipage}{0.195\textwidth}
% \centering
% \includegraphics[height=1\textwidth]{images/pfd2cs/graphs/1920.png}
% \includegraphics[width=1\textwidth]{images/pfd2cs/layouts/1920.png}
% \includegraphics[width=1\textwidth]{images/pfd2cs/images/1920.png}
% \end{minipage}
\begin{minipage}{0.195\textwidth}
\centering
\includegraphics[height=1\textwidth]{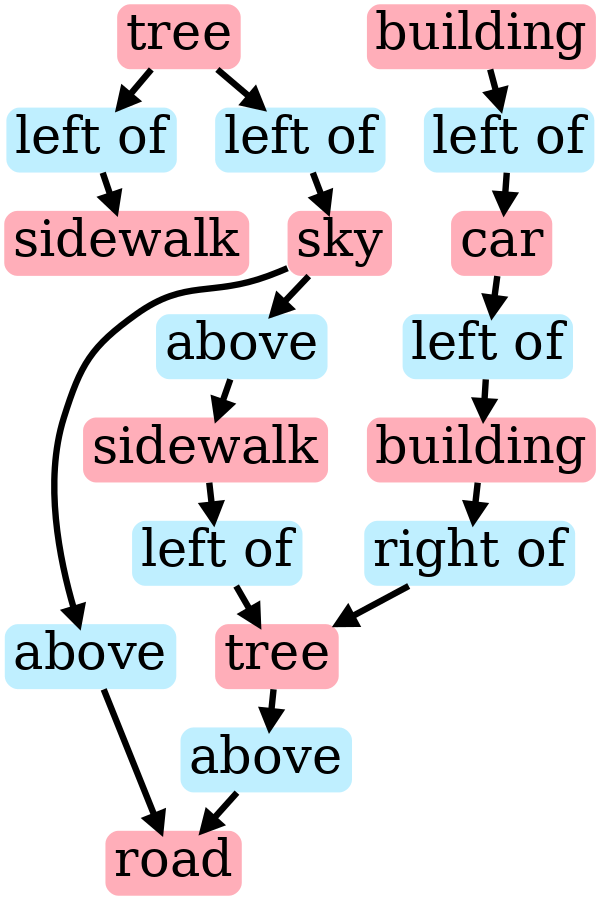}
\includegraphics[width=1\textwidth]{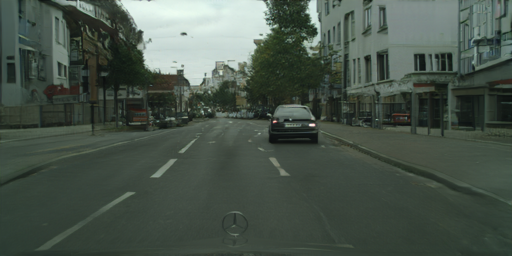}
\includegraphics[width=1\textwidth]{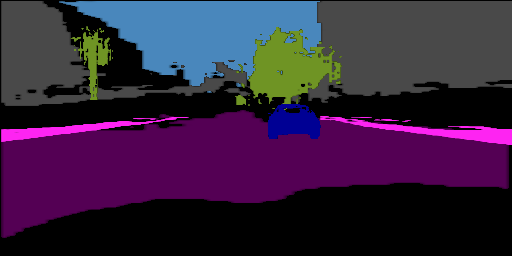}
\end{minipage}

\caption{Examples of synthetic scene graphs and corresponding generated traffic scenes from Cityscapes.}
\label{fig:results_cityscapes}
\end{figure*}
\begin{figure*}[ht]

% \begin{minipage}{0.195\textwidth}
% \centering
% \includegraphics[height=1\textwidth]{images/pfd2bdd/graphs/980.png}
% \includegraphics[width=1\textwidth]{images/pfd2bdd/images/980.png}
% \end{minipage}
% \vspace{0.005\textwidth}
\begin{minipage}{0.195\textwidth}
\centering
\includegraphics[height=1\textwidth]{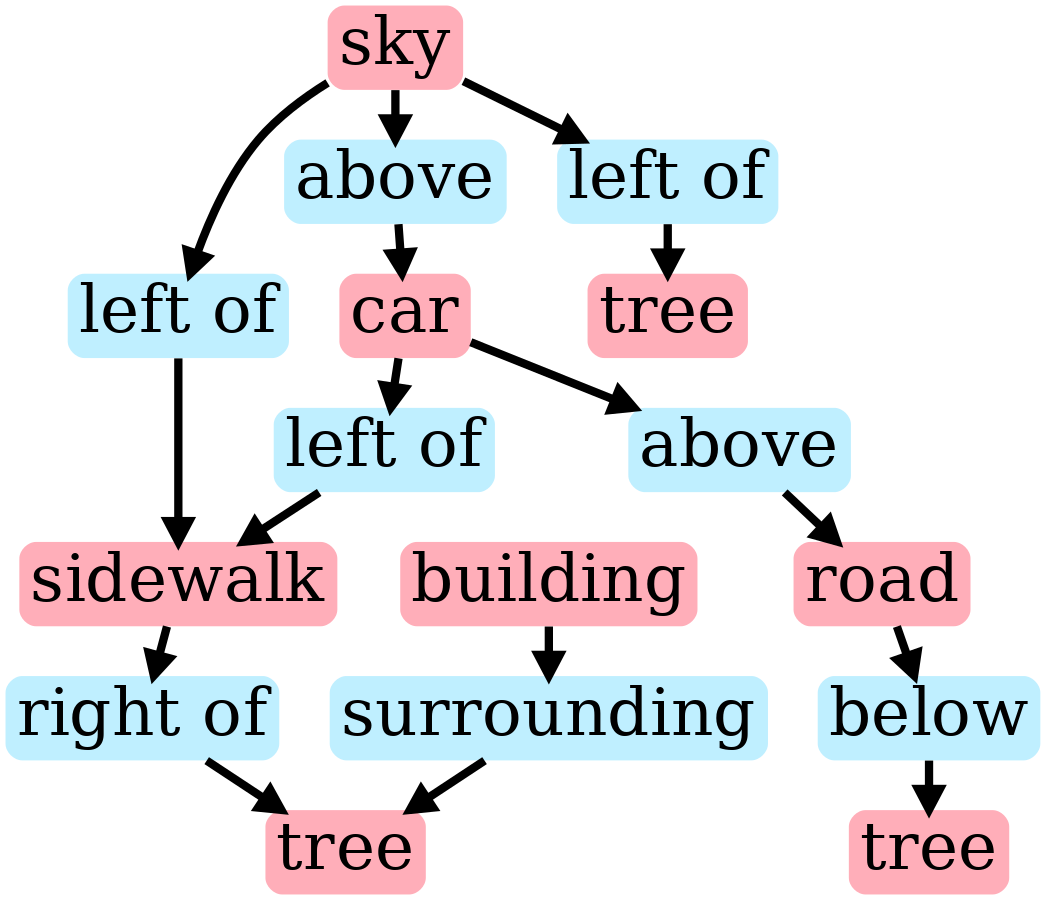}
\includegraphics[width=1\textwidth]{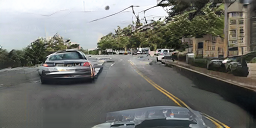}
\includegraphics[width=1\textwidth]{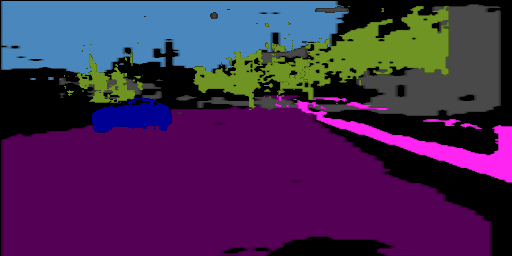}
\end{minipage}
\vspace{0.005\textwidth}
\begin{minipage}{0.195\textwidth}
\centering
\includegraphics[height=1\textwidth]{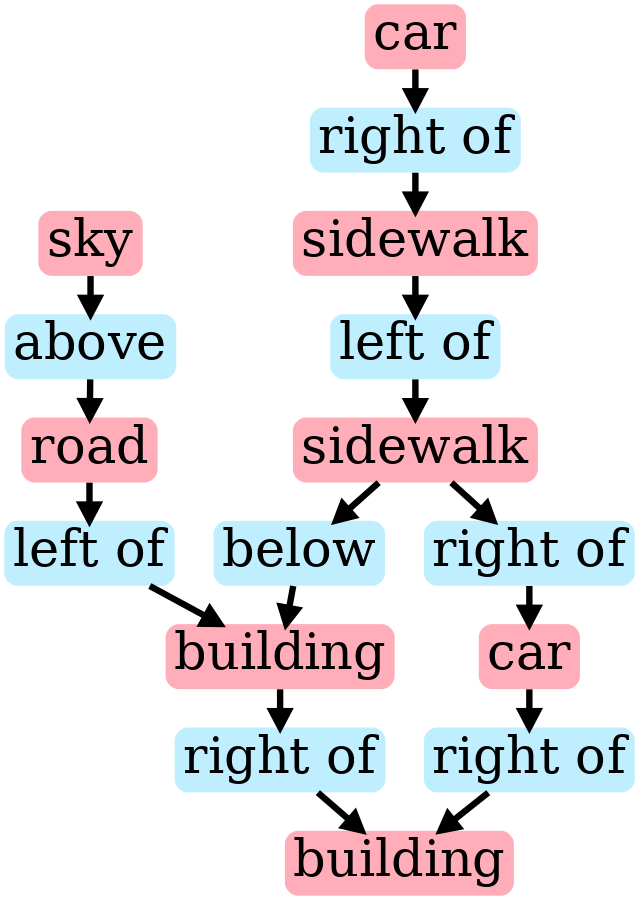}
\includegraphics[width=1\textwidth]{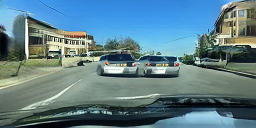}
\includegraphics[width=1\textwidth]{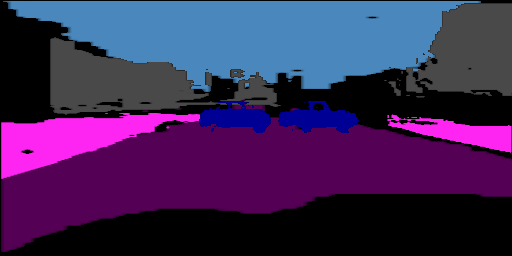}
\end{minipage}
\vspace{0.005\textwidth}
\begin{minipage}{0.195\textwidth}
\centering
\includegraphics[height=1\textwidth]{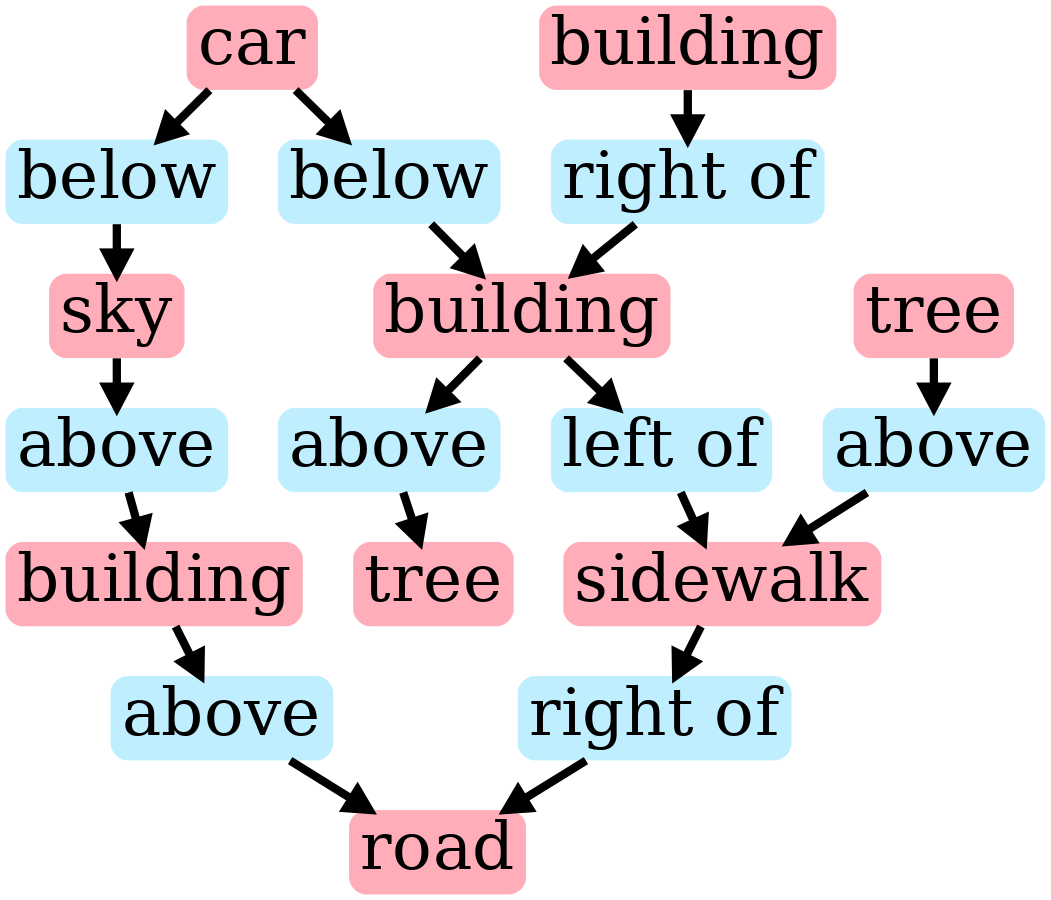}
\includegraphics[width=1\textwidth]{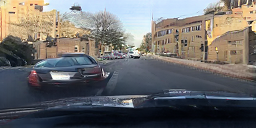}
\includegraphics[width=1\textwidth]{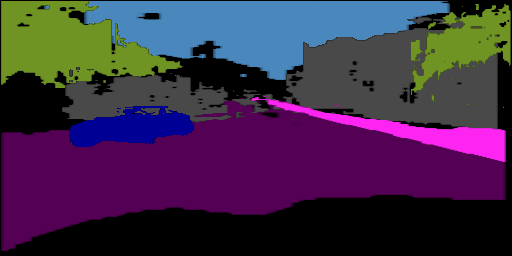}
\end{minipage}
\vspace{0.005\textwidth}
\begin{minipage}{0.195\textwidth}
\centering
\includegraphics[height=1\textwidth]{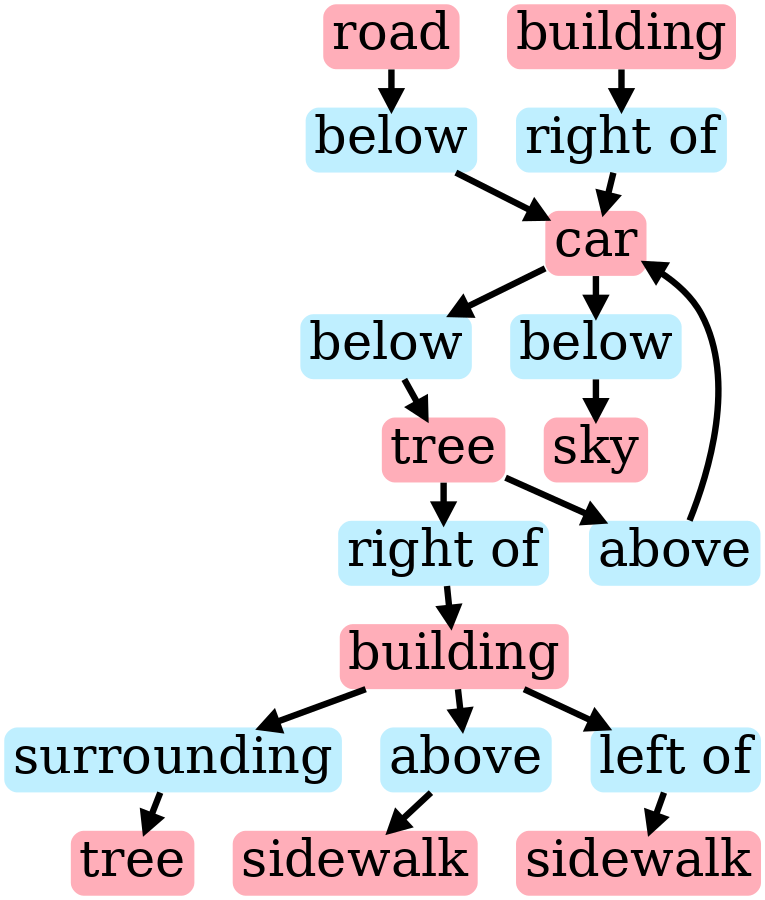}
\includegraphics[width=1\textwidth]{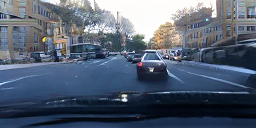}
\includegraphics[width=1\textwidth]{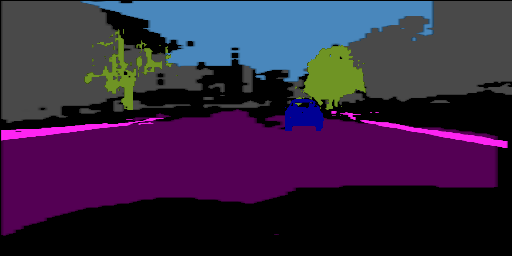}
\end{minipage}
\vspace{0.005\textwidth}
\begin{minipage}{0.195\textwidth}
\centering
\includegraphics[height=1\textwidth]{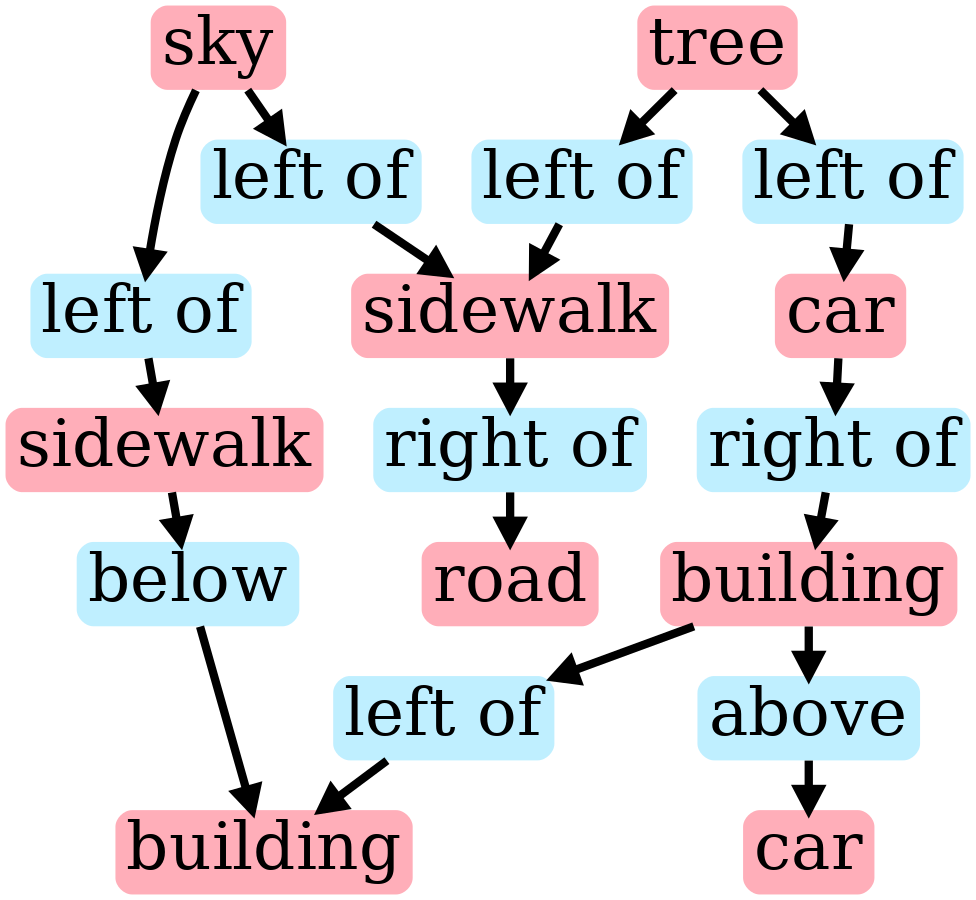}
\includegraphics[width=1\textwidth]{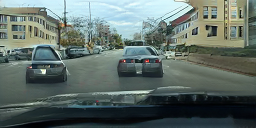}
\includegraphics[width=1\textwidth]{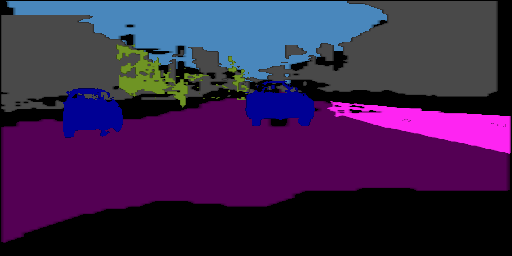}
\end{minipage}
\vspace{0.005\textwidth}
\caption{Examples of synthetic scene graphs with corresponding BDD generated traffic scenes and semantic maps.}
\label{fig:results_bdd}
\end{figure*}
\section{Proposed Approach}
Our approach is focused on synthetic scene graphs and unsupervised traffic scene generation, where pairs of synthetic and real scenes are not available. We derive a synthetic scene graph from a procedurally generated scene that does not provide visual characteristics such as textures or materials. For the synthesized scene graph, we then produce an image of the corresponding traffic scene, which resembles the content of the underlying synthetic scene and the realistic appearance of the target data. Figure~\ref{fig:scheme} provides an overview of our method. It highlights the two main parts of the approach: synthetic scene graph generation and realistic image generation.

\subsection{Synthetic 3D Scene Graphs}
We adopt the setup from \cite{Johnson2018} for traffic scene data, where every scene graph is represented by nodes $n_i$ associated to each object $i$ in the scene and edges $e_{ij}$ encode relations between particular nodes $n_i$ and $n_j$. Our graph processor is built upon \cite{Ashual2019} but extends it in several ways, allowing 3D scene graph construction. In addition to the simplified relations - \textit{e.g.}, \textit{left of}, \textit{above} \cite{Ashual2019}, we integrate 3D information about the scene in form of spatial relations (\textit{in front of}, \textit{behind})  between objects and spatial attributes (depth component \textit{z}) of particular objects. Importantly, this information is available in simulation at no cost. To generate a comprehensive traffic scene, we extend the objects list and integrate background classes such as \textit{sky, building, vegetation}.

\begin{figure*}[t!]
\begin{minipage}{0.325\textwidth}
\centering
\includegraphics[height=0.5\textwidth]{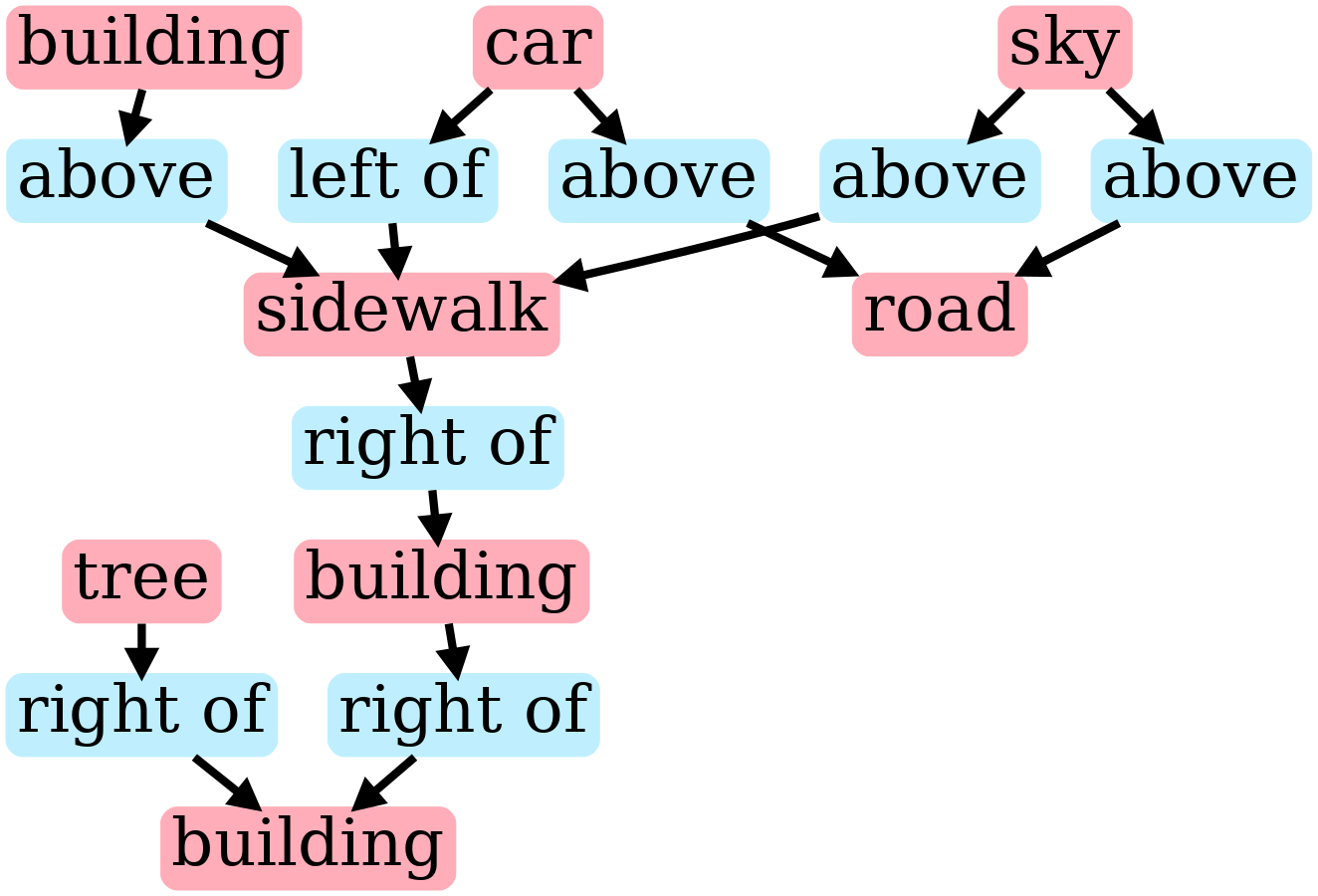}
\includegraphics[height=0.5\textwidth]{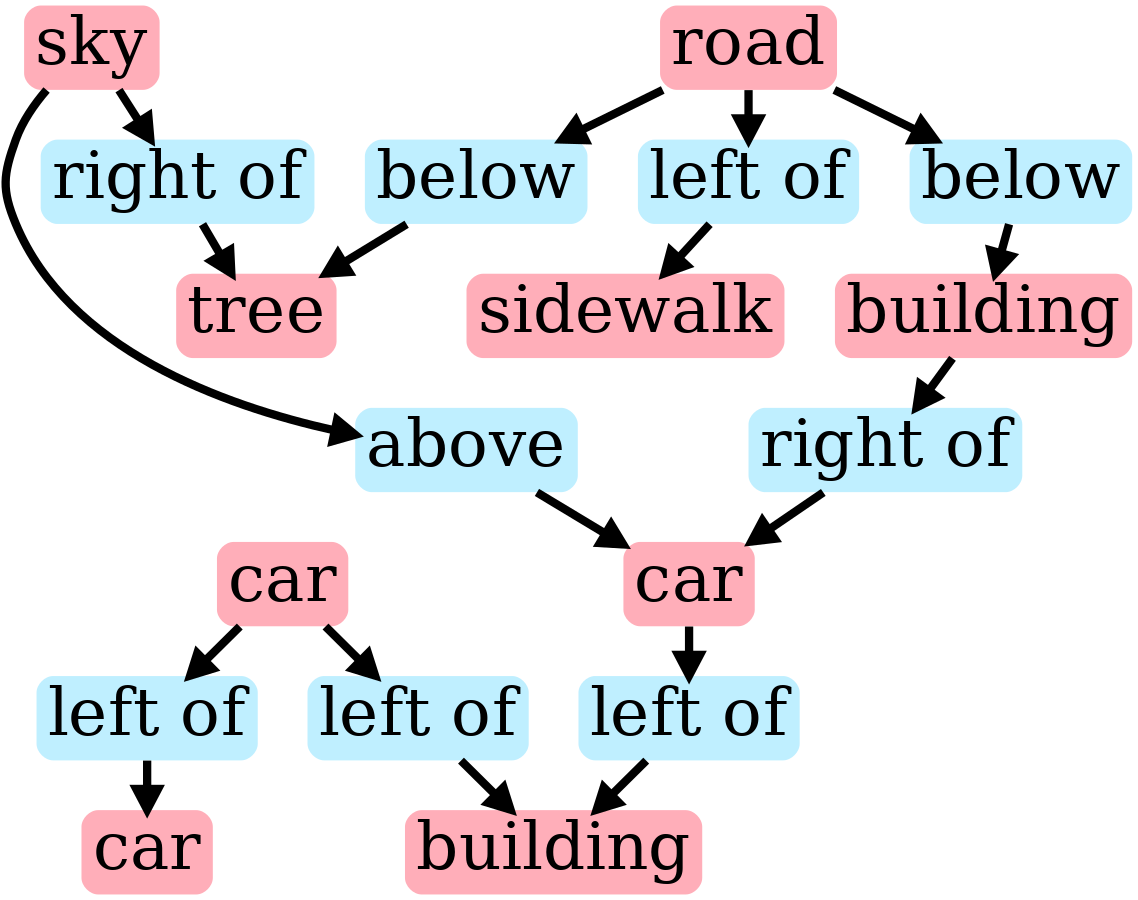}
% \caption*{Synthetic Scene Graph}
\end{minipage}
\begin{minipage}{0.325\textwidth}
\includegraphics[width=1\textwidth]{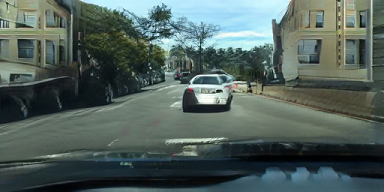}
\includegraphics[width=1\textwidth]{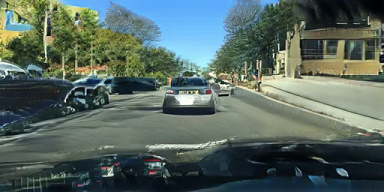}
% \caption*{Generated BDD Traffic Scene}
\end{minipage}
\begin{minipage}{0.325\textwidth}
\includegraphics[width=1\textwidth]{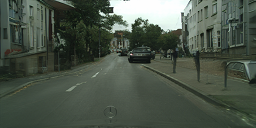}
\includegraphics[width=1\textwidth]{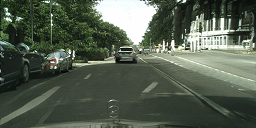}
% \caption*{Generated Cityscapes Traffic Scene}
\end{minipage}
\caption{BDD and Cityscapes traffic scenes generated from the same synthetic scene graph.}
\label{fig:joint_cs_bdd}
\end{figure*}

Therefore, every node in our graph is represented as $ n_i = [o_i, l_i, z_i], o_i \in \mathbb{R}, l_i \in \{0,1\}^{L\times L}, z_i \in \{0,1\}^{Z}$, where $o_i$ takes one of the indexes of $C$ possible object classes, $L$ is size of the grid to denote objects location $l_i$ and $Z$ is the scene depth to denote objects position $z_i$ along the z-axis. Analogously, every edge is represented as $e_{i,j} \in \mathbb{R}$, taking one of the values from predefined dictionary of relation types.

To train the graph processor, we also obtain mask $m_i$ and the bounding box $b_i \in \mathbb{R}^{4}$ for each object $i$ from the simulation. Hence, our graph processor $P$ is trained in a supervised manner to produce the full scene layout $t \in \mathbb{R}^{H \times\ W \times C}$. Similar to \cite{Johnson2018}, our graph processor $P$ is composed of 3 networks: \textit{graph network}, \textit{mask regression network} and \textit{box regression network}. The graph network is constructed of graph convolutional layers which extract features from scene graphs and encode per-object layout embedding, mask network consists of several deconvolution layers, and box network is a Multi-Layer Perceptron (MLP). The last two networks are dedicated to predicting masks and bounding boxes of the objects present in the scene.

As an output of the first step, we obtain a synthetic scene graph alongside the corresponding layout $t$ produced by the graph processor $P$. Both scene graphs and layouts are domain agnostic, therefore, we omit visual information from the synthetic domain (rendering) and merely encode content.

\subsection{Traffic Scenes}
In the next step, we apply the image generator $G$ to layout $t$ to produce a realistic traffic scene image $x$, which visually resembles the target data $X=\{\hat{x} \in \mathcal{X}\}$ while at the same time preserving the content of the original scene graph. Since our approach focuses on the realistic image generation from the synthetic scene graphs, it requires unsupervised training, as synthetic-real pairs are not available. To enable unsupervised training we employ adversarial \cite{Goodfellow2014} and contrastive \cite{Gutmann2010} losses. Hence, our final objective is as follows:
\begin{align}
\begin{split}
&\mathcal{L} = \mathcal{L}_{SG} + \mathcal{L}_{TS} \\
&= \mathcal{L}_{MSE}(b_i,\hat{b}_i) + \mathcal{L}_{GAN}(m_i,\hat{m}_i) + \mathcal{L}_{FM}(m_i,\hat{m}_i)\\
& + \mathcal{L}_{GAN}(x,\hat{x}) + \mathcal{L}_{NCE}(x,\hat{x})\\
\end{split}
\label{eq:loss}
\end{align}
Where $\mathcal{L}_{FM}$ is a \textit{feature matching loss} \cite{Salimans2016}, $\mathcal{L}_{NCE}$ is a \textit{multilayer, patch-wise contrastive loss} \cite{Chen2020, Park2020} and $\mathcal{L}_{FM}$ is an \textit{adversarial loss} applied both to masks $m_i$ and images $x$. With $\hat{}$ we denote either a synthetic ground-truth (for masks) or a target dataset (for images):
\begin{align}
\begin{split}
&\mathcal{L}_{GAN}(m_i,\hat{m}_i) = \\
&\mathbb{E} \log D_{m}(m_i) + \mathbb{E} \log (1-D_{m}(\hat{m}_i))\\
&\mathcal{L}_{FM}(m_i,\hat{m}_i) = \norm{\mathbb{E}f(m_i) - \mathbb{E}_{x \sim}f(\hat{m}_i)}_2^2 \\
&\mathcal{L}_{GAN}(x,\hat{x}) = \mathbb{E}_{x\sim X} \log D(x) + \mathbb{E}_{\hat{x}\sim X} \log (1-D(\hat{x}))\\
&\mathcal{L}_{NCE}(x,\hat{x}) = \mathbb{E}_{x\sim X} \sum_{l} \sum_{s} \ell(\hat{x}_l^s, x_l^s, \overline{x}_l^s)
\end{split}
\label{eq:losses}
\end{align}
Where $f$ denotes activations on an intermediate layer of the discriminator, $\ell$ is a cross-entropy loss function for a positive pair of examples and $x_l^s$ are the features of the generator from $l$-th layer at $s$-th location \cite{Chen2020}.

\section{Experiments}
To properly evaluate our approach, we design the synthetic graph generator to operate on bounding boxes and semantic as input. This allows producing synthetic scene graphs for any existing synthetic image dataset, which provides 3D bounding boxes and semantic segmentation labels. In particular we use 2 synthetic traffic datasets for scene graph generation: PfB \cite{Richter2017} and Synscapes \cite{Wrenninge2020}. Both are large-scale datasets that provide about 25000 urban traffic environment images at $1914\times1052$ and $1440\times720$ alongside 2D/3D bounding boxes, semantic and instance maps.

In the image generation part, we utilize two real datasets, Cityscapes \cite{Cordts2016} and Berkley Deep Drive \cite{Yu2018} as providers of appearance characteristics. The first one comprises about 3000 images of traffic scenes from several German cities with finely annotated semantic maps. The second one includes about 100000 images of the streets of US American cities, of which we only use 10000.

\begin{figure*}[t!]
\centering
\begin{minipage}{0.325\textwidth}
\includegraphics[width=1\textwidth]{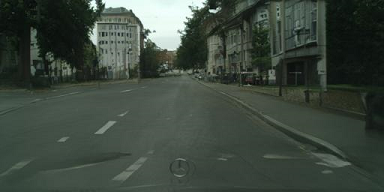}
\caption*{No car}
\end{minipage}
\begin{minipage}{0.325\textwidth}
\includegraphics[width=1\textwidth]{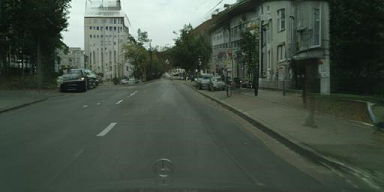}
\caption*{\textit{z}-attribute 5}
\end{minipage}
\begin{minipage}{0.325\textwidth}
\includegraphics[width=1\textwidth]{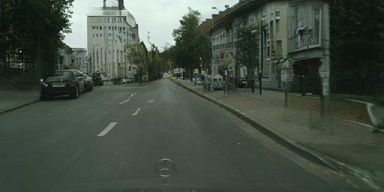}
\caption*{\textit{z}-attribute 2}
\end{minipage}
\caption{An example of traffic scene manipulation by changing spatial attribute of the car object.}
\label{fig:results_z}
\end{figure*}
\begin{figure*}[t!]
    \centering
\begin{minipage}{0.325\textwidth}
\includegraphics[width=1\textwidth]{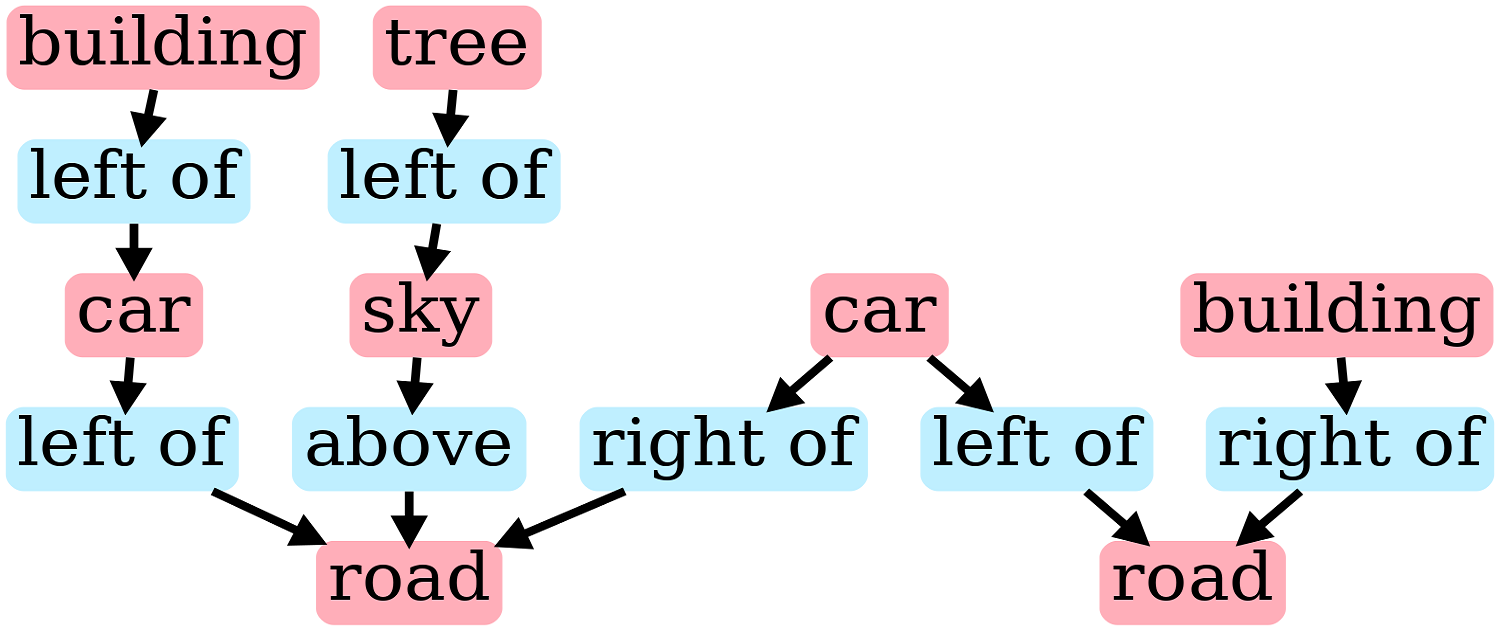}
\includegraphics[width=1\textwidth]{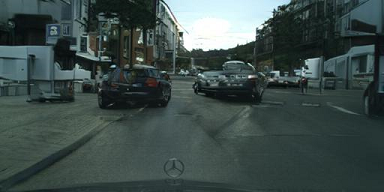}
\caption*{No relation}
\end{minipage}
\begin{minipage}{0.325\textwidth}
\includegraphics[width=1\textwidth]{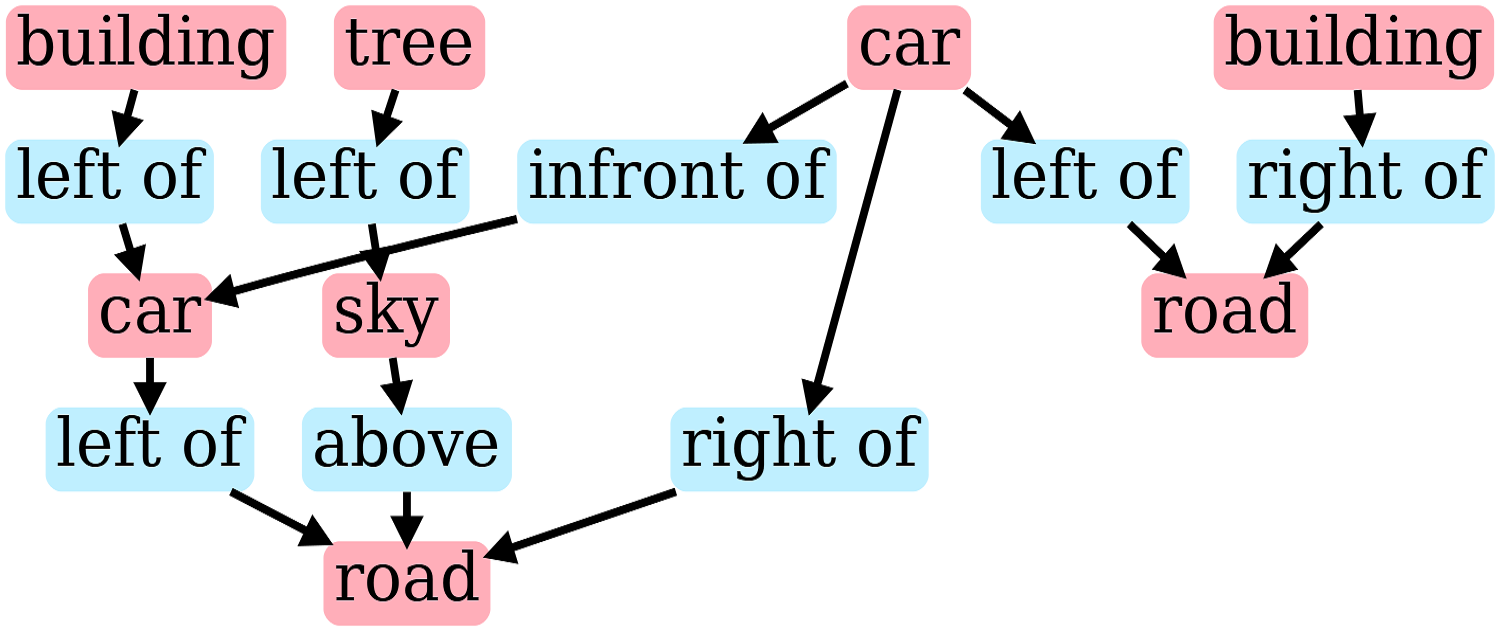}
\includegraphics[width=1\textwidth]{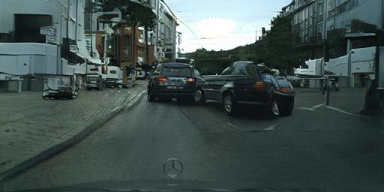}
\caption*{\textit{in front of}}
\end{minipage}
\begin{minipage}{0.325\textwidth}
\includegraphics[width=1\textwidth]{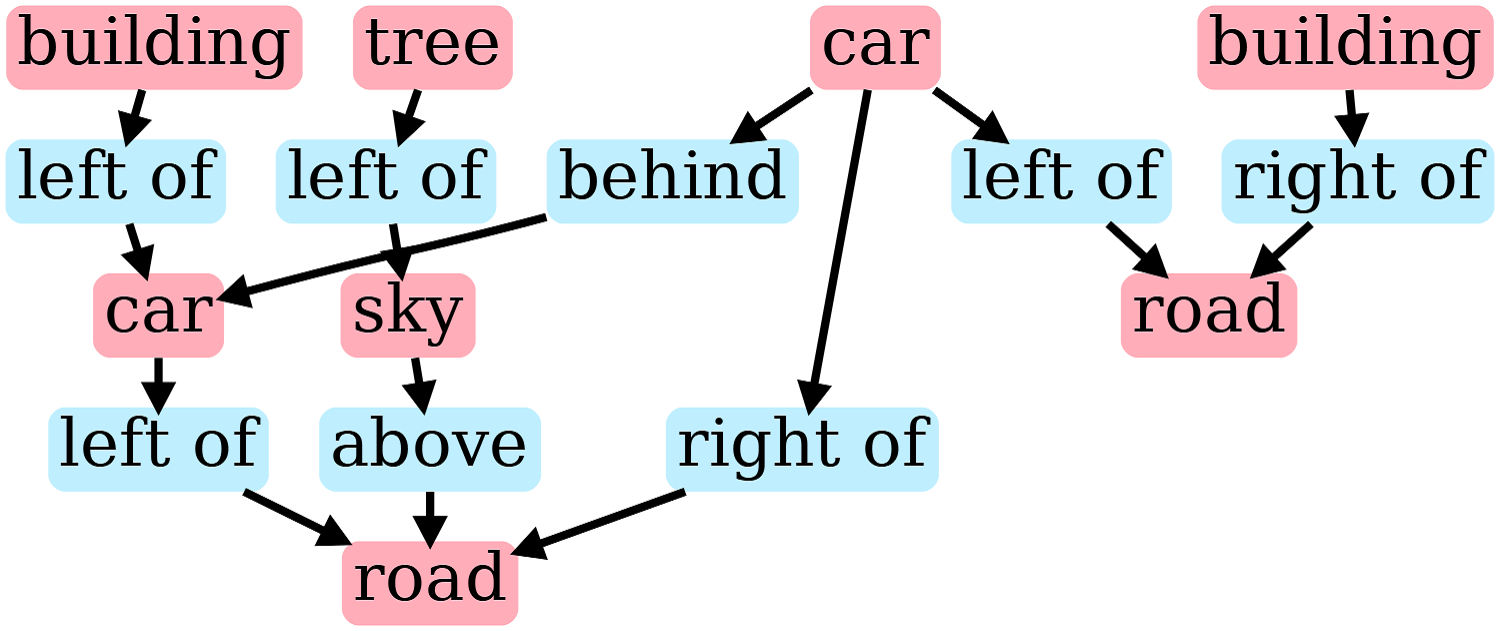}
\includegraphics[width=1\textwidth]{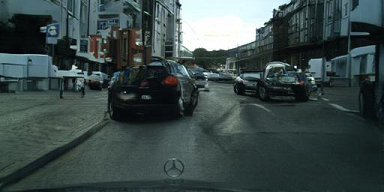}
\caption*{\textit{behind}}
\end{minipage}
\caption{An example of traffic scene manipulation by changing spatial relation of 2 car objects.}
\label{fig:behind_before}
\end{figure*}
\begin{figure*}[t!]
    \centering
\begin{minipage}{0.245\textwidth}
\includegraphics[width=1\textwidth]{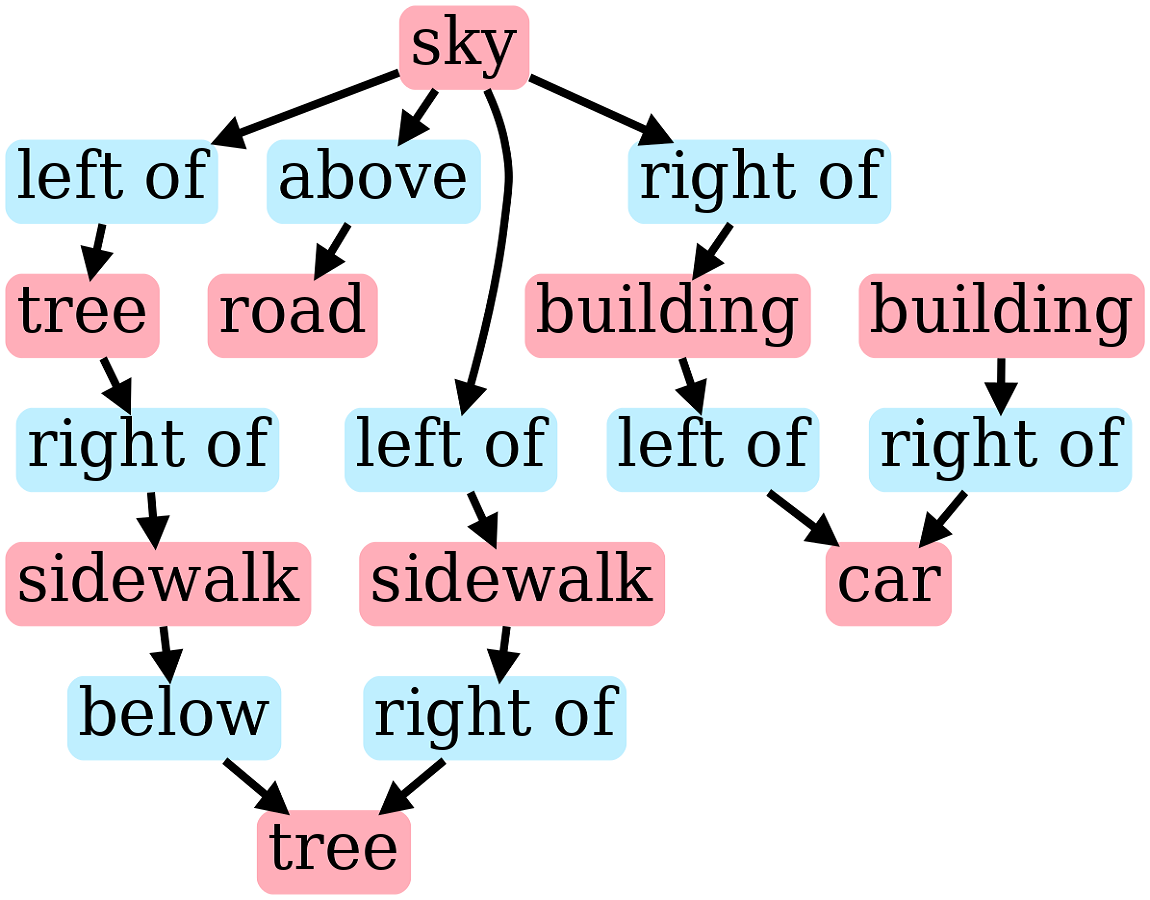}
\includegraphics[width=1\textwidth]{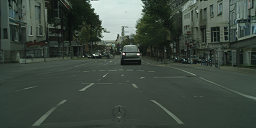}
\caption*{\textit{Building right of the car}}
\end{minipage}
\begin{minipage}{0.245\textwidth}
\includegraphics[width=1\textwidth]{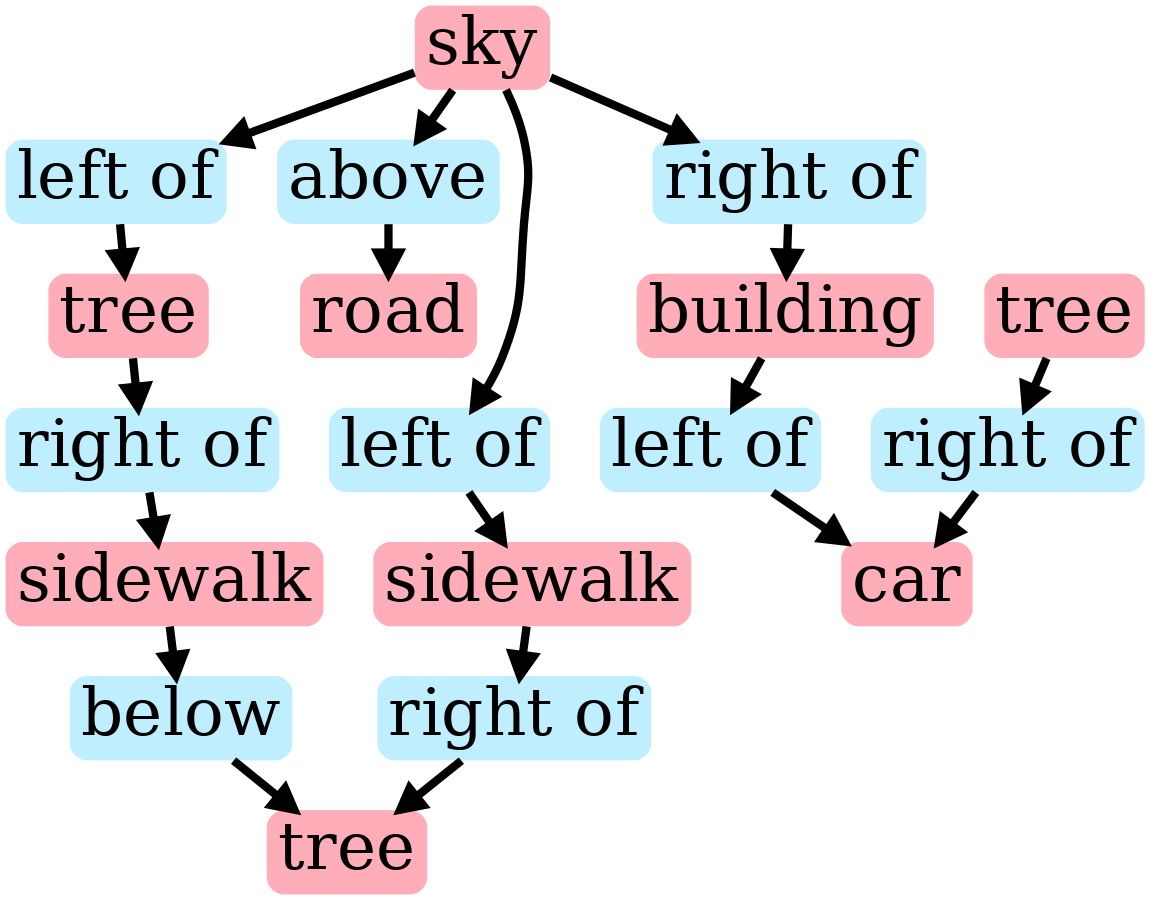}
\includegraphics[width=1\textwidth]{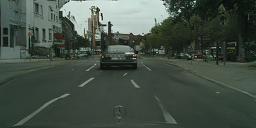}
\caption*{\textit{Vegetation right of the car}}
\end{minipage}
\begin{minipage}{0.245\textwidth}
\includegraphics[width=1\textwidth]{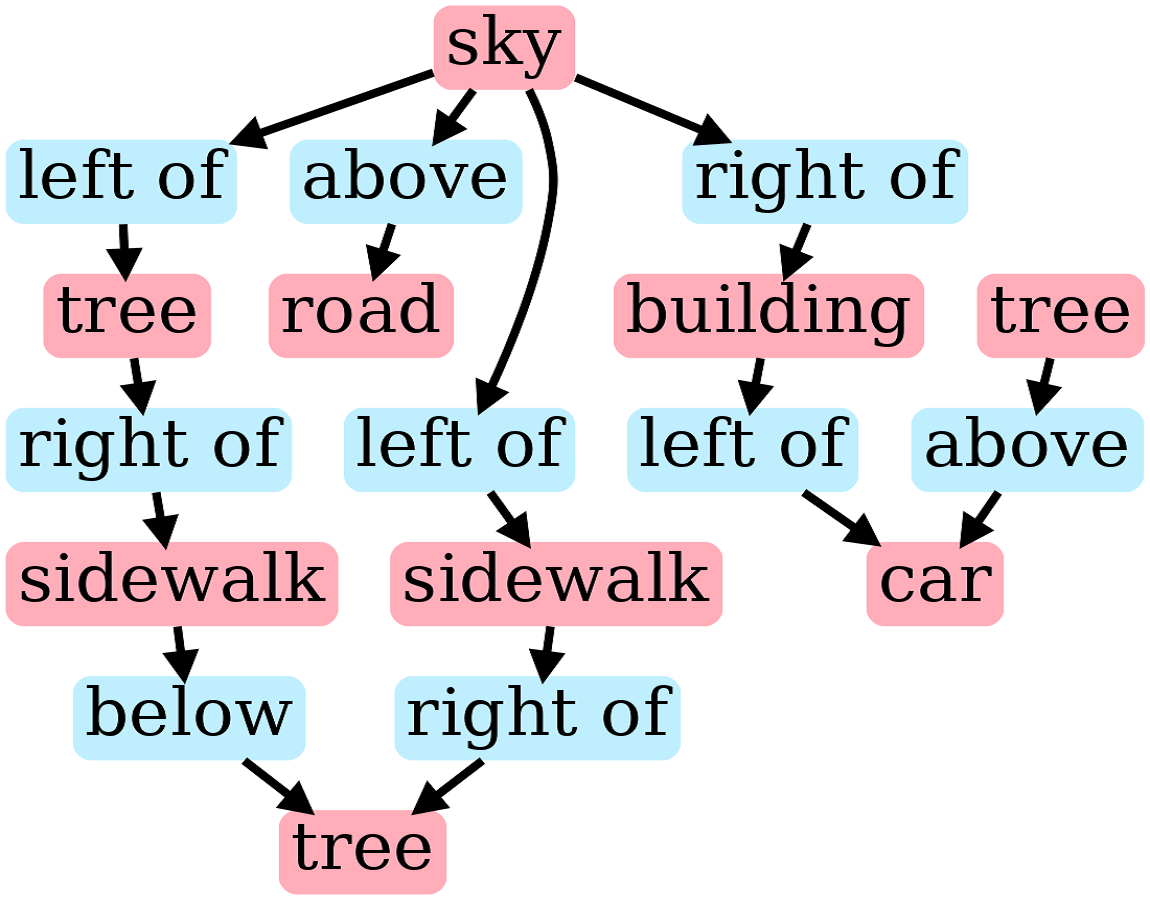}
\includegraphics[width=1\textwidth]{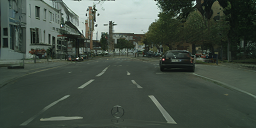}
\caption*{\textit{Vegetation above the car}}
\end{minipage}
\begin{minipage}{0.245\textwidth}
\includegraphics[width=1\textwidth]{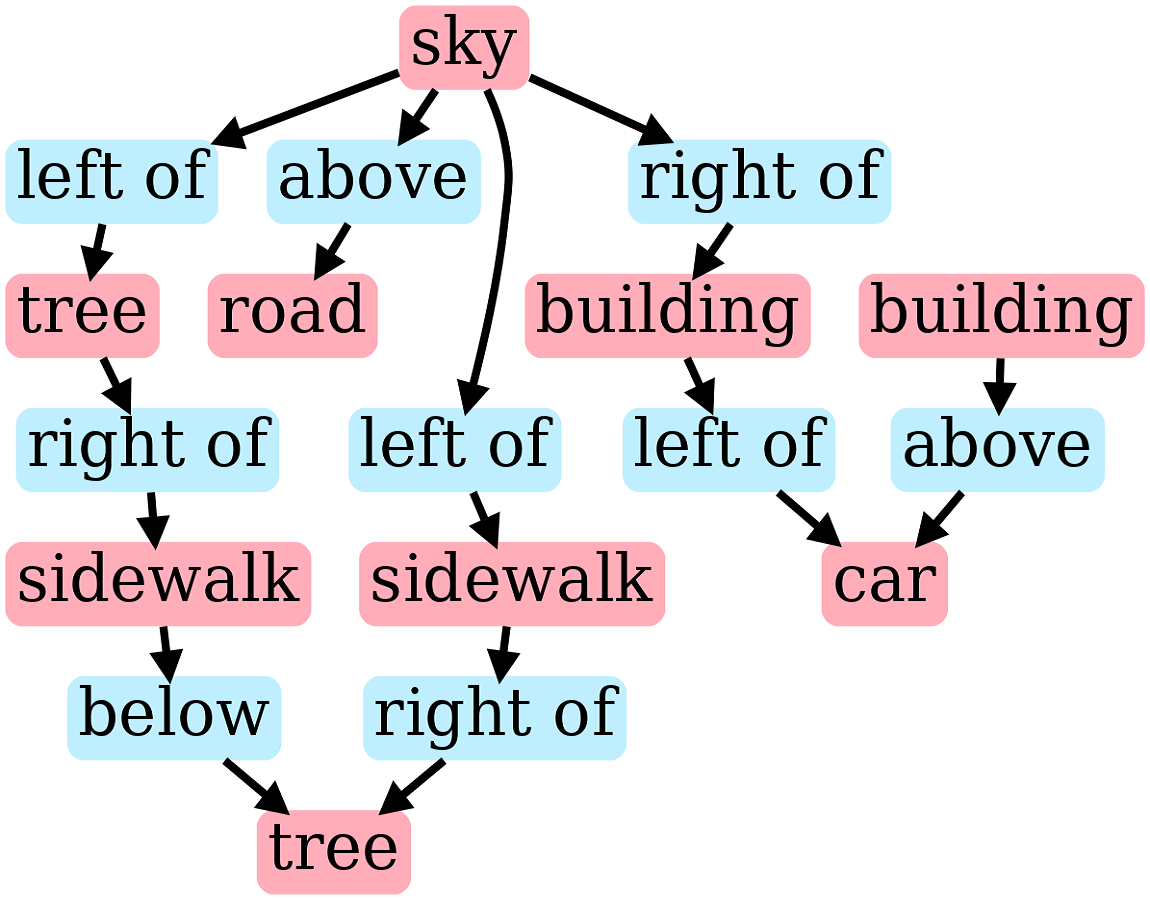}
\includegraphics[width=1\textwidth]{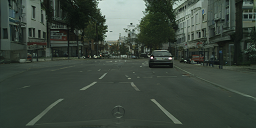}
\caption*{\textit{Building above the car}}
\end{minipage}
\caption{An example of traffic scene manipulation by changing classes.}
\label{fig:manipulation_classes}
\end{figure*}
\begin{figure*}[t!]
\centering
\includegraphics[width=0.95\textwidth]{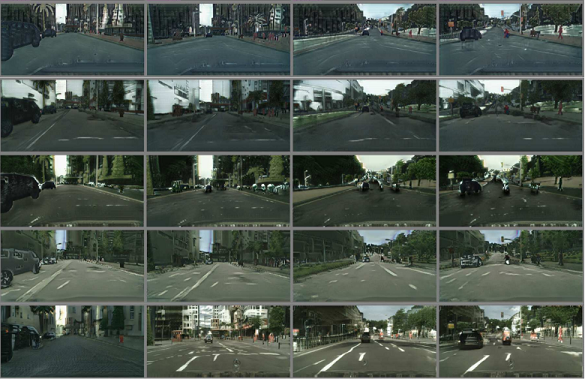}
\caption{Comparison of the generated images (top$\rightarrow{}$bottom): CycleGAN\cite{Zhu2017}, DualGAN\cite{Yi2017}, MUNIT\cite{Huang2018}, DRIT\cite{Shu2018}, ours}\label{fig:comparison}
\end{figure*}

\subsection{Synthesis}
We first produce a set of synthetic scene graphs that includes 5k samples. We rely on this set to train our scene graph generator. For both experiments with Cityscapes and BDD, our setup is similar. We conduct all experiments at the resolution of $256x\times512$ pixels. The training continues for 150 epochs with a learning rate of $1e-3$ and momentum $0.99$. Figures \ref{fig:results_cityscapes} and \ref{fig:results_bdd} show the results for the training on Cityscapes and BDD, respectively. Additionally, in figure~\ref{fig:comparison} we compare our method with state-of-the-art unsupervised image translation techniques - \textit{e.g.}, CycleGAN \cite{Zhu2017}, MUNIT \cite{Huang2018}. This is also reflected in the FID score - CycleGAN achieves 103.05, MUNIT - 75.98 and our method 47.26.

It is evident from the aforementioned figures that the traffic scene generator picks up the appearance of both datasets nicely, producing in general realistic images. Being slightly inaccurate in fine details (especially those of cars), it is yet very consistent on the class and image level. The generator produces a very congruous appearance of the objects of following classes: \textit{road, sidewalk, building, vegetation}.
We also want to highlight the conditioning of the generation process illustrated in Figures \ref{fig:results_cityscapes} and \ref{fig:results_bdd} in sections \ref{sec:experiments_classes} and \ref{sec:experiments_spatial}. There we demonstrate traffic scene manipulation with regard to newly introduced classes, spatial attributes and spatial relations to ensure that our improvements take effect.

\subsection{Spatial Attributes and Relations}\label{sec:experiments_spatial}
Synthetic scene graphs let us leverage spatial information about the traffic scene. Figure \ref{fig:results_z} visualizes 3 traffic scenes generated from the same scene graph by varying the z-component of a particular object in the graph. In the first image, no car is present in the scene. In the following images, we put a car in the scene and change its position by manipulating the value of the \textit{z}-attribute.

The 3D information for the scene not only enables changing the z-attribute of the objects, but also enables spatial relation between them. Thus, Figure \ref{fig:behind_before} shows a traffic scene produced by the scene graph by swapping the relation between two cars, which appear in the scene having approximately the same size when not bounded by any spatial relation. The right car gets pulled toward the ego vehicle, while the left car appears farther away from the case of \textit{in front of} relation. When the relation changes to \textit{behind}, the left car is rendered significantly bigger than before and bigger than the right car.

\begin{table*}[t]
\begin{center}
\resizebox{\textwidth}{!}{
\begin{tabular}{l|cccccccc|c}
\hline
& Sky & Road & Tree & Building & Person & Car & Bus & Truck & Mean\\
\hline
PfB 25k \cite{Richter2017} & 62.8 & 41.1 & 67.8 & 64.3 & 14.0 & 38.8 & 1.1 & 8.1 & 37.3 \\
Ours 3D 5k & 23.4 & 76.1 & 44.7 & 46.3 & 8.1 & 41.2 & 3.5 & 1.6 & 30.6 \\
Ours 3D 10k & 34.2 & 68.0 & 62.5 & 60.8 & 9.7 & 40.7 & 1.3 & 3.0 & 35.0 \\
Ours 3D 5k + PfB & 25.4 & 80.2 & 55.9 & 64.6 & 29.3 & 64.2 & 6.3 & 7.7 & 41.7 \\
Ours 3D 2k* & 64.7 & 91.3 & 67.4 & 64.9 & 13.2 & 39.5 & 1.5 & 2.8 & 43.2 \\
\hline
\end{tabular}}    
\end{center}
\caption{Segmentation results of DRN26 network. Training performed on generated data and evaluation made on Cityscapes~\textit{val} \cite{Cordts2016} dataset. (*) denotes class balanced scene graphs.}
\label{tab:iou_unsupervised}
\end{table*}

\subsection{Traffic Scene Classes}\label{sec:experiments_classes}

Additionally to objects (\textit{car, bus}) we introduce multiple classes which are characteristic for traffic scene scenarios. Such classes include \textit{road, sidewalk, vegetation, building, sky}. To verify that introducing background classes is effective, we perform several experiments by manipulating the classes in the traffic scenes. Figure \ref{fig:manipulation_classes} shows the effects of such manipulation: changing classes of the particular nodes in the scene graph results in the respective adjustment of the semantic layout of the generated scene.

In addition to, to a qualitative evaluation of the approach, we conduct several experiments on the downstream task of semantic segmentation. This provides a quantitative assessment of the data generated by our method.

\subsection{Image Segmentation}
To assess the proposed method quantitatively, we train a state-of-the-art semantic segmentation method on the data produced by our method as well as on the underlying synthetic data. Here we focus on the scene graphs produced from the Cityscapes simulation - \textit{i.e}, we randomly generate 2 sets of scene graphs with 5000 and 10000 samples. For these scene graphs, we generate corresponding images alongside semantic maps. They are then used to train DRN \cite{Yu2016}.

We train a segmentation network for 200 epochs on 8 classes of interest and evaluate on the real Cityscapes \textit{val} dataset. We provide per class IoU score and meanIoU also referred as \textit{Jaccard Index} \cite{Everingham2015} in Table~\ref{tab:iou_unsupervised}. The table demonstrates that DRN trained on the generated 5k dataset lies 5\% behind DRN trained on the original simulated PFD 25k dataset. Doubling the number of scene graphs and corresponding images reduces the gap to 2 points. Additionally, we report in the table~\ref{tab:iou_unsupervised} the results of the experiment where scene graphs follow the statistics of a real dataset. Knowing the target data class ratio allows us to sub-sample the data and keep those synthetic scene graphs whose layouts follow target data class ratio. This makes it possible to reduce the scene graph number to 2000 samples and improve segmentation performance by 5\% compared to the original 25k dataset.

\section{Conclusion}
In this work, we propose a method to ease data generation for realistic traffic scenes from domain agnostic scene representation called scene graphs instead of using photo-realistic rendering. We utilize synthetic scene graphs enhanced by spatial attributes (\textit{z}) and spatial relations (e.g., \textit{behind}). Furthermore, we introduce an unsupervised approach for realistic image generation from synthetic scene graphs. The approach shows convincing generation results as demonstrated in the proposed qualitative evaluation. We also show the effectiveness of our method through traffic scene manipulation and validation on a downstream task.

\section{Acknowledgment}
The research leading to these results is funded by the German Federal Ministry for Economic Affairs and Energy within the project “KI Absicherung – Safe AI for Automated Driving". The authors would like to thank the consortium for the successful cooperation.
% \newpage
{
\bibliographystyle{ieee}
\bibliography{literature}

\begin{thebibliography}{10}\itemsep=-1pt

\bibitem{Almahairi2018}
A.~Almahairi, S.~Rajeshwar, A.~Sordoni, P.~Bachman, and A.~Courville.
\newblock Augmented {C}ycle{GAN}: Learning many-to-many mappings from unpaired
  data.
\newblock In {\em International Conference on Machine Learning}, 2018.

\bibitem{Ashual2019}
O.~Ashual and L.~Wolf.
\newblock Specifying object attributes and relations in interactive scene
  generation.
\newblock In {\em IEEE International Conference on Computer Vision (ICCV)},
  2019.

\bibitem{Brock2018}
A.~Brock, J.~Donahue, and K.~Simonyan.
\newblock Large scale {GAN} training for high fidelity natural image synthesis.
\newblock In {\em International Conference on Learning Representations}, 2019.

\bibitem{Cabon2020}
Y.~Cabon, N.~Murray, and M.~Humenberger.
\newblock Virtual kitti 2, 2020.

\bibitem{Chen2017}
Q.~Chen and V.~Koltun.
\newblock Photographic image synthesis with cascaded refinement networks.
\newblock In {\em IEEE International Conference on Computer Vision (ICCV)},
  2017.

\bibitem{Chen2020}
T.~Chen, S.~Kornblith, M.~Norouzi, and G.~Hinton.
\newblock A simple framework for contrastive learning of visual
  representations.
\newblock In {\em International Conference on Machine Learning}, 2020.

\bibitem{Cordts2016}
M.~Cordts, M.~Omran, S.~Ramos, T.~Rehfeld, M.~Enzweiler, R.~Benenson,
  U.~Franke, S.~Roth, and B.~Schiele.
\newblock The cityscapes dataset for semantic urban scene understanding.
\newblock In {\em IEEE CVPR}, 2016.

\bibitem{Dhamo2020}
H.~Dhamo, A.~Farshad, I.~Laina, N.~Navab, G.~D. Hager, F.~Tombari, and
  C.~Rupprecht.
\newblock Semantic image manipulation using scene graphs.
\newblock In {\em IEEE CVPR}, 2020.

\bibitem{Dosovitskiy2017}
A.~Dosovitskiy, G.~Ros, F.~Codevilla, A.~Lopez, and V.~Koltun.
\newblock {CARLA}: {An} open urban driving simulator.
\newblock In {\em Annual Conference on Robot Learning}, 2017.

\bibitem{Everingham2015}
M.~Everingham, S.~M.~A. Eslami, L.~Van~Gool, C.~K.~I. Williams, J.~Winn, and
  A.~Zisserman.
\newblock The pascal visual object classes challenge: A retrospective.
\newblock {\em International Journal of Computer Vision}, 2015.

\bibitem{Gaidon2016}
A.~Gaidon, Q.~Wang, Y.~Cabon, and E.~Vig.
\newblock Virtual worlds as proxy for multi-object tracking analysis.
\newblock In {\em IEEE conference on Computer Vision and Pattern Recognition},
  2016.

\bibitem{Goodfellow2014}
I.~Goodfellow, J.~Pouget-Abadie, M.~Mirza, B.~Xu, D.~Warde-Farley, S.~Ozair,
  A.~Courville, and Y.~Bengio.
\newblock Generative adversarial nets.
\newblock In {\em Advances in Neural Information Processing Systems}, 2014.

\bibitem{Gulrajani2017}
I.~Gulrajani, F.~Ahmed, M.~Arjovsky, V.~Dumoulin, and A.~C. Courville.
\newblock Improved training of wasserstein gans.
\newblock In {\em Advances in Neural Information Processing Systems}, 2017.

\bibitem{Gutmann2010}
M.~Gutmann and A.~Hyvärinen.
\newblock Noise-contrastive estimation: A new estimation principle for
  unnormalized statistical models.
\newblock In {\em International Conference on Artificial Intelligence and
  Statistics}, 2010.

\bibitem{Huang2018}
X.~Huang, M.-Y. Liu, S.~Belongie, and J.~Kautz.
\newblock Multimodal unsupervised image-to-image translation.
\newblock In {\em ECCV}, 2018.

\bibitem{Isola2017}
P.~Isola, J.-Y. Zhu, T.~Zhou, and A.~A. Efros.
\newblock Image-to-image translation with conditional adversarial networks.
\newblock In {\em IEEE Conference on Computer Vision and Pattern Recognition
  (CVPR)}, 2017.

\bibitem{Johnson2018}
J.~Johnson, A.~Gupta, and L.~Fei-Fei.
\newblock Image generation from scene graphs.
\newblock In {\em IEEE CVPR}, 2018.

\bibitem{Johnson2015}
J.~Johnson, R.~Krishna, M.~Stark, L.-J. Li, D.~Shamma, M.~Bernstein, and
  L.~Fei-Fei.
\newblock Image retrieval using scene graphs.
\newblock In {\em IEEE Conference on Computer Vision and Pattern Recognition},
  2015.

\bibitem{Karras2019}
T.~Karras, S.~Laine, M.~Aittala, J.~Hellsten, J.~Lehtinen, and T.~Aila.
\newblock Analyzing and improving the image quality of stylegan.
\newblock In {\em CVPR}, 2020.

\bibitem{Krishna2017}
R.~Krishna, Y.~Zhu, O.~Groth, J.~Johnson, K.~Hata, J.~Kravitz, S.~Chen,
  Y.~Kalantidis, L.-J. Li, D.~A. Shamma, M.~S. Bernstein, and L.~Fei-Fei.
\newblock Visual genome: Connecting language and vision using crowdsourced
  dense image annotations.
\newblock {\em IJCV}, 2017.

\bibitem{McCormac2017}
J.~McCormac, A.~Handa, S.~Leutenegger, and A.~J. Davison.
\newblock Scenenet rgb-d: Can 5m synthetic images beat generic imagenet
  pre-training on indoor segmentation?
\newblock In {\em ICCV}, 2017.

\bibitem{Newell2017}
A.~Newell and J.~Deng.
\newblock Pixels to graphs by associative embedding.
\newblock In {\em Advances in Neural Information Processing Systems}, 2017.

\bibitem{Odena2017}
A.~Odena, C.~Olah, and J.~Shlens.
\newblock Conditional image synthesis with auxiliary classifier {GAN}s.
\newblock In {\em International Conference on Machine Learning}, 2017.

\bibitem{Park2020}
T.~Park, A.~A. Efros, R.~Zhang, and J.-Y. Zhu.
\newblock Contrastive learning for unpaired image-to-image translation.
\newblock In {\em European Conference on Computer Vision}, 2020.

\bibitem{Park2019}
T.~Park, M.-Y. Liu, T.-C. Wang, and J.-Y. Zhu.
\newblock Semantic image synthesis with spatially-adaptive normalization.
\newblock In {\em IEEE Conference on Computer Vision and Pattern Recognition},
  2019.

\bibitem{Qiu2017}
W.~Qiu, F.~Zhong, Y.~Zhang, S.~Qiao, Z.~Xiao, T.~S. Kim, Y.~Wang, and
  A.~Yuille.
\newblock Unrealcv: Virtual worlds for computer vision.
\newblock {\em ACM Multimedia Open Source Software Competition}, 2017.

\bibitem{Richter2017}
S.~R. Richter, Z.~Hayder, and V.~Koltun.
\newblock Playing for benchmarks.
\newblock In {\em IEEE International Conference on Computer Vision}, 2017.

\bibitem{Ros2016}
G.~Ros, L.~Sellart, J.~Materzynska, D.~Vazquez, and A.~M. Lopez.
\newblock The synthia dataset: A large collection of synthetic images for
  semantic segmentation of urban scenes.
\newblock In {\em CVPR}, 2016.

\bibitem{Salimans2016}
T.~Salimans, I.~Goodfellow, W.~Zaremba, V.~Cheung, A.~Radford, and X.~Chen.
\newblock Improved techniques for training gans.
\newblock In {\em NIPS}, 2016.

\bibitem{Shah2017}
S.~Shah, D.~Dey, C.~Lovett, and A.~Kapoor.
\newblock Airsim: High-fidelity visual and physical simulation for autonomous
  vehicles.
\newblock In {\em Field and Service Robotics}, 2017.

\bibitem{Shu2018}
R.~Shu, H.~H. Bui, H.~Narui, and S.~Ermon.
\newblock A dirt-t approach to unsupervised domain adaptation.
\newblock In {\em ICLR}, 2018.

\bibitem{Sugiyama2012}
M.~Sugiyama and M.~Kawanabe.
\newblock {\em Machine Learning in Non-Stationary Environments: Introduction to
  Covariate Shift Adaptation}.
\newblock The MIT Press, 2012.

\bibitem{Wang2018}
T.-C. Wang, M.-Y. Liu, J.-Y. Zhu, A.~Tao, J.~Kautz, and B.~Catanzaro.
\newblock High-resolution image synthesis and semantic manipulation with
  conditional gans.
\newblock In {\em IEEE/CVF CVPR}, 2018.

\bibitem{Wrenninge2020}
M.~Wrenninge and J.~Unger.
\newblock Synscapes: A photorealistic synthetic dataset for street scene
  parsing.
\newblock In {\em https://arxiv.org/abs/1810.08705}, 2020.

\bibitem{Yi2017}
Z.~Yi, H.~Zhang, P.~Tan, and M.~Gong.
\newblock Dualgan: Unsupervised dual learning for image-to-image translation.
\newblock In {\em {ICCV}}, 2017.

\bibitem{Yu2018}
F.~Yu, H.~Chen, X.~Wang, W.~Xian, Y.~Chen, F.~Liu, V.~Madhavan, and T.~Darrell.
\newblock Bdd100k: A diverse driving dataset for heterogeneous multitask
  learning.
\newblock In {\em https://arxiv.org/abs/1810.08705}, 2018.

\bibitem{Yu2016}
F.~Yu and V.~Koltun.
\newblock Multi-scale context aggregation by dilated convolutions.
\newblock In {\em International Conference on Learning Representations (ICLR)},
  2016.

\bibitem{Zhang2017}
H.~Zhang, T.~Xu, H.~Li, S.~Zhang, X.~Wang, X.~Huang, and D.~Metaxas.
\newblock Stackgan: Text to photo-realistic image synthesis with stacked
  generative adversarial networks.
\newblock In {\em {ICCV}}, 2017.

\bibitem{Zhu2017}
J.-Y. Zhu, T.~Park, P.~Isola, and A.~A. Efros.
\newblock Unpaired image-to-image translation using cycle-consistent
  adversarial networks.
\newblock In {\em IEEE International Conference on Computer Vision (ICCV)},
  2017.

\end{thebibliography}
}

% \section{Appendix}
% \input{figure_cs_bdd_h}

\end{document}